\documentclass{article}

\usepackage[final]{neurips_2021}
\usepackage{xr-hyper}
\makeatletter
\newcommand*{\addFileDependency}[1]{
  \typeout{(#1)}
  \@addtofilelist{#1}
  \IfFileExists{#1}{}{\typeout{No file #1.}}
}
\makeatother

\newcommand*{\myexternaldocument}[1]{%
    \externaldocument{#1}%
    \addFileDependency{#1.tex}%
    \addFileDependency{#1.aux}%
}
\myexternaldocument{neurips_2021-suppl}

\usepackage[utf8]{inputenc} 
\usepackage[T1]{fontenc}    
\usepackage[dvipsnames]{xcolor}
\usepackage[colorlinks=true,linkcolor=Blue,citecolor=Blue]{hyperref}       
\usepackage{url}            
\usepackage{booktabs}       
\usepackage{amsfonts}       
\usepackage{nicefrac}       
\usepackage{microtype}      
\usepackage{amsmath}
\usepackage{amsthm}
\usepackage{enumitem}
\newtheorem{theorem}{Theorem}

\newtheorem{definition}{Definition}

\newtheorem{example}{Example}

\usepackage{subfig}
\usepackage{multirow}
\usepackage{bbm}
\newcommand{\ie}[0]{\textit{i.e.}}
\newcommand{\eg}[0]{\textit{e.g.}}
\usepackage[noabbrev]{cleveref}

\usepackage{graphicx}
\usepackage{bm}
\usepackage{bbm}
\usepackage{arydshln}
\newcommand{\bmm}[1]{\bm{#1}}
\newcommand{\indep}{\perp \!\!\!\!\! \perp}
\newcommand{\dep}{ \not\!\perp\!\!\!\!\!\perp}

\newcommand{\Imp}{\text{Imp}}
\usepackage{ulem}

\title{From global to local MDI variable importances for random forests and when they are Shapley values}

\author{%
  Antonio Sutera\thanks{Corresponding author. Email address: \href{mailto:sutera.antonio@gmail.com}{sutera.antonio@gmail.com}},\, Gilles Louppe,\, Van Anh Huynh-Thu,\, Louis Wehenkel,\, Pierre Geurts\\
  Dept. of EE \& CS, University of Li\`{e}ge, Belgium\\
  \texttt{\{a.sutera,g.louppe,vahuynh,l.wehenkel,p.geurts\}@uliege.be}
}

\begin{document}

\maketitle
\begin{abstract}
Random forests have been widely used for their ability to provide so-called \textit{importance measures}, which give insight at a global (per dataset) level on the relevance of input variables to predict a certain output. On the other hand, methods based on Shapley values have been introduced to refine the analysis of feature relevance in tree-based models to a local (per instance) level. In this context, we first show that the global Mean Decrease of Impurity (MDI) variable importance scores correspond to Shapley values under some conditions. Then, we derive a \textit{local} MDI importance measure of variable relevance, which has a very natural connection with the global MDI measure and can be related to a new notion of local feature relevance. We further link local MDI importances with Shapley values and discuss them in the light of related measures from the literature. The measures are illustrated through experiments on several classification problems.

\end{abstract}

\section{Motivation}

While research in machine learning (ML) often focuses on predictive accuracy, another important topic concerns the interpretation of ML models and their predictions. Interpreting a model helps to uncover the mechanisms it captures (\eg, biomarkers useful to diagnose a disease), and to explain its predictions (\eg, why a particular patient is diagnosed healthy or sick). The latter becomes essential when a ML prediction may impact one's life, and as a way of checking the trustworthiness of a ML model (\eg, to identify unwanted biases). 
Common interpretation tools include variable importance measures that assess which, and to which extent, variables are important for a model. They help to understand how the model works, and to gain insight on the underlying modelled mechanism.

In tree-based methods, such as Random forests \citep{breiman2001random}, feature importance scores can be derived as a low-cost by-product of the learning step. 
Given their extensive use in applied research, tree-based importance measures have been studied both empirically (see, \eg, \citet{strobl2007bias,archer2008empirical,genuer2010variable,auret2011empirical}) and theoretically (see, \eg, \citet{ishwaran2007variable,louppe2013understanding,louppe2014understanding,sutera2018random,li2019debiased,sutera2019importance,scornet2020trees}). Assuming a sufficiently large learning set and number of trees, these works showed that importance measures have desirable properties, such as consistency with respect to the notion of feature relevance. They also analyzed the impact of learning meta-parameters (\eg, randomization level, tree depth, ensemble size) on such properties.
While standard importance measures evaluate the \textit{global} importance of a feature at the level of a dataset, several works proposed new approaches based on Shapley values to derive \textit{local} scores reflecting the importance of a feature for a given prediction (\eg, \citet{neto2020explainable,lundberg2020local,izza2020explaining}). 

The contribution of the present work in this context is two-fold. First, we show that the standard mean decrease of impurity (MDI) measure when derived from totally randomized trees and in asymptotic conditions (similar to those used to show the consistency with respect to the relevance) are Shapley values, and therefore have the same properties as any other importance measures based on these values (Section \ref{sec:global+shapley}). Secondly, we propose a new local MDI measure for tree-based ensemble models to measure feature relevance locally, which naturally derives from global MDI and corresponds to Shapley values in the same conditions (Section \ref{sec:local+shapley}). Global and local MDI are compared against other Shapley-value based scores, both conceptually (Section \ref{sec:relatedwork}) and empirically (Section \ref{sec:illustration}).

\section{Background}\label{sec:background}

In what follows, we consider a standard supervised learning setting and denote by $V=\{X_1,\dots,X_p\}$ the set of $p$ input variables, and by $Y$ the output.

\paragraph{Game theory and Shapley value.}
We only remind here concepts and results that are useful later in the paper. Notations below are mostly adapted from \citep{besner2019axiomatizations}.

In game theory, a \textit{TU-game} $(V,v)$ (\ie, \textit{a cooperative game} with transferable utilities) is defined by a finite set of players $V=\{X_1,\dots,X_p\}$\footnote{We use the same notations for the (set of) players as for (set of) input features, as the two will coincide later.} and a \textit{characteristic (or coalition) function} $v\in\mathbb{V}:2^V\rightarrow\mathbb{R}$, with $v(\emptyset)=0$, that maps a \textit{coalition} (\ie, a set) of players to a real number representing the outcome or gain of a game (see, \eg, \citep{van2015proper}). A TU-game $(V,v)$ is {\it monotonic} if $v(S)\leq v(T)$ if $S\subseteq T\subseteq V$. Let us denote by $V^{-m}$ the set $V\setminus \{X_m\}$. The marginal contribution $MC_m^v(S)$ of player $X_m \in V$ for $S\subseteq V^{-m}$ is defined by $MC_m^v(S) = v(S\cup \{X_m\}) - v(S)$. A player $X_m\in V$ is called a {\it null player} if $MC_m^v(S) = 0$ for all $S\subseteq V^{-m}$. Two players $X_i$ and $X_j$ are said to be {\it symmetric} whenever $v(S\cup\{X_i\}) = v(S\cup\{X_j\})$ for all coalitions $S\subseteq V^{-i,j}$.

A \textit{TU-value} $\varphi_v: V\rightarrow \mathbb{R}$ is a function that assigns to any player $X_m\in V$ and any function $v$ ($\in \mathbb{V}$ is omitted in the rest) a value, denoted $\varphi_v(X_m) \in \mathbb{R}$, also known as its \textit{payoff}, reflecting its contribution in the game $(V,v)$. Several properties or axioms for TU-values have been defined in the literature that are expected to be satisfied in practical contexts (see \citep{besner2019axiomatizations} for a more exhaustive list):

    \textbf{Efficiency:} For all $v$
    , $\sum_{X_m\in V} \varphi_v(X_m)=v(V)$. {\it The TU-value divides the total gain (\ie, gain when {\normalfont all} players are involved) among all players in an additive way.}
    
    \textbf{Symmetry:} For all $v$, 
    $\varphi_v(X_i)=\varphi_v(X_j)$ if players $X_i$ and $X_j$ are symmetric. {\it Two players of equal contributions in every game (\ie, with every coalition $S$) should get the same value.}
    
    \textbf{Null player:} For all $v$
    , $\varphi_v(X_m)=0$ if $X_m$ is a null player. {\it A null player should get a zero payoff.}
    
    \textbf{Strong monotonocity\footnote{In the ML literature, strong monotonicity is often called consistency \citep{lundberg2017unified}.}:} For all $v,w$ 
    and $X_m\in V$ such that $MC_m^v(S) \geq MC_m^w(S)$ for all $S\subseteq V^{-m}$, we have $\varphi_v(X_m)\ge \varphi_w(X_m)$. {\it If a player's marginal contributions are greater (or equal) in a game than in another in all coalitions, then its payoff in this game should not be lower than in the other.}

Specific forms of TU-value have been studied in the literature from the point of view of which axioms they satisfy and how uniquely they are defined by these axioms. As one of the most prominent results, it has been shown \citep{young1985monotonic} that the only TU-value that satisfies Efficiency, Symmetry, Null player\footnote{Actually, the Null player property is not required as it can be derived from strong monotonicity.}, and Strong monotonicity is the {\it Shapley value} $\phi^{Sh}_v$ defined by \citep{shapley1953value}:
\begin{eqnarray}\label{eq:shapleyvaldef}
\phi^{Sh}_v(X_m) 
=\sum_{S\subseteq V^{-m}} \dfrac{|S|!(p-|S|-1)!}{p!} MC_m^v(S).
\end{eqnarray}
Other equivalent formulations of the Shapley value, as well as other axiomatisations of this value, have been proposed in the literature. Axiomatisations of other, typically more general, sets of TU-values are also available (see \citep{besner2019axiomatizations} for a recent and exhaustive discussion of this topic).

\paragraph{Feature relevance.}\label{sec:background:relevance}

In the feature selection literature, a common definition of the relevance of a feature is as follows \citep{kohavi1997wrappers}:

\textit{    A variable $X_m\in V$ is {\bf relevant} to $Y$ (with respect to $V$) iff $\exists B\subset V: X_m\dep Y|B$. A variable $X_m$ is {\bf irrelevant} if it is not relevant.}
Relevant variables can be further divided into two categories according to their degree of relevance \citep{kohavi1997wrappers}:
\textit{    A variable $X_m$ is {\bf strongly} relevant to $Y$ (with respect to $V$) iff $Y\dep X_m|V^{-m}$. A variable X is {\bf weakly} relevant if it is relevant but not strongly relevant.}
Strongly relevant variables thus convey information about the output that no other variable (or combination of variables) in $V$ conveys. 

\paragraph{Decision trees and forests.}

Each interior node of a decision tree \citep{breiman1984classification} is labelled with a test based on some input and each leaf node is labelled with a value of the output. The tree is typically grown  from a learning sample of size $N$ drawn from $P(V,Y)$ using a procedure that recursively partitions the samples at each node $t$ into two child nodes ($t_L$ and $t_R$). The test $s_t$ used to partition the samples at node $t$ is the one that maximises the mean decrease of some node impurity measure $i(\cdot)$ (\eg, the Shannon entropy, the Gini index or the variance of $Y$): $\Delta i(s,t) = i(t) - \tfrac{p(t_L)}{p(t)} i(t_L) - \tfrac{p(t_R)}{p(t)} i(t_R),$
where $p(t_L)$ and $p(t_R)$ are the proportions of samples that fall in nodes $t_L$ and $t_R$ respectively. Single decision trees suffer from a high variance that is very efficiently reduced by building instead an ensemble of randomized trees and aggregating their predictions. Popular methods are \citet{breiman2001random}'s Random Forests that build each tree from a different bootstrap sample with a local random selection of $K(\le p)$ variables at each node from which to identify the best split, and \cite{geurts2006extremely}'s Extra-Trees which skip bootstrapping and additionally randomly select the split values. {Following \citet{geurts2006extremely,louppe2013understanding}, ensemble of randomized trees grown with the value of the randomization parameters $K$ set to 1 will be called {\it Totally randomized trees}.}

\paragraph{Mean decrease impurity importance.}\label{sec:mdi}

Given an ensemble of trees, several methods have been proposed to evaluate the (global) importance of variables for predicting the output \citep{breiman1984classification, breiman2001random}.  This paper focuses on the Mean Decrease of Impurity (MDI) importance. Given a tree $T$, the MDI importance of a variable $X_m$ for predicting the output $Y$ is defined as : 
\begin{equation}\label{eqn:mdi_tree}
\Imp(X_m,T) =\sum_{t\in T: \nu(s_t)=X_m} p(t) \Delta i(s_t,t),
\end{equation}
where  the sum is over all interior nodes $t$ in $T$, $\nu(s_t)$ denotes the variable tested at node $t$, and $p(t)$ is the fraction of samples reaching node $t$. $\Imp(X_m,T)$ is thus the (weighted) sum of impurity decreases over all nodes where $X_m$ is used to split.
The MDI importance of $X_m$ derived from forests of $N_T$ trees is then the average of $\Imp(X_m,T)$ over all trees: 
\begin{equation} 
\Imp(X_m) = \dfrac{1}{N_T}\sum_{T}  \Imp(X_m,T). \end{equation}
While this measure was initially proposed as a heuristic, \cite{louppe2013understanding} characterise it theoretically under the following conditions: (1) all input variables and the output are categorical (not necessarily binary) (2) trees use so-called \textit{exhaustive splits}\footnote{Each node is split into $|\mathcal{X}_i|$ sub-trees, one for each of the  $|\mathcal{X}_i|$ different values of the split variable $X_i$.}, and (3) impurity is measured by Shannon entropy\footnote{A short introduction to information theory and the related notations used in the paper is provided in Appendix G.}. Later, we will refer to these conditions collectively as the {\it categorical setting}.

In the categorical setting, \cite{louppe2013understanding} (Thm. 1) show that for totally randomized trees (\ie, $K=1$) and in \textit{asymptotic conditions} (\ie, assuming $N_T\rightarrow
\infty$ and a learning sample of infinite size), the MDI importance, denoted $\Imp_{\infty}$, is given by:
\begin{equation}\label{eqn:popMDI}
    \Imp_\infty(X_m) = \sum_{k=0}^{p-1}\dfrac{1}{C_p^k}\dfrac{1}{p-k} \sum_{B\in\mathcal{P}_k(V^{-m})}I(Y;X_m|B),
\end{equation}
where $\mathcal{P}_k(V^{-m})$ is the set of subsets of $V^{-m}$ of cardinality $k$, and $I(Y;X_m|B)$ is the conditional mutual information of $X_m$ and $Y$ given the variables in $B$. They also show that the sum of the MDI importances of all variables is equal to the mutual information between all input features and the output \citep[Thm. 2]{louppe2013understanding}:
\begin{equation}\label{eq:sumimp}
\sum_{m=1}^p \Imp_{\infty}(X_m)=I(Y;V).
\end{equation}
A direct consequence of Equation \ref{eqn:popMDI} is that a variable $X_m$ is irrelevant iff $\Imp_\infty(X_m)=0$, which makes the MDI importance a sensible measure to identify relevant variables.

\section{\textit{Global} MDI importances are Shapley values}\label{sec:global+shapley}

In this section, we revisit the theoretical analysis of \citet{louppe2013understanding} and \citet{sutera2019importance} in the light of TU-games and TU-values, focusing on the categorical setting and asymptotic conditions adopted by these authors. We show in Section \ref{sec:globalmdishapley} that MDI importances computed from totally randomized trees can be interpreted as the Shapley value for a particular TU-game and we then discuss  in Section~\ref{sec:globalmdikgt1} the case of non totally randomized trees (\ie, $K>1$).

\subsection{Totally randomized trees}\label{sec:globalmdishapley}

Let us consider a TU-game $(V,v)$, where $V$ is the set of variables and the
coalition function $v$ is the mutual information $v(\cdot)=I(Y;\cdot)$. Since
$v(\emptyset) = I(Y;\emptyset) = 0$, this is a valid TU-game. This TU-game is monotonic
since we have $I(Y;T) = I(Y;S) + I(Y;T\setminus S|S)\geq I(Y;S)$ as soon as
$S\subseteq T\subseteq V$ (using the chain rule and the positivity of the 
conditional mutual information). Marginal contributions for $v$ can be rewritten as:
\begin{equation}
MC_m^v(S) = v(S\cup \{X_m\}) - v(S) = I(Y;S\cup\{X_m\}) - I(Y;S) = I(Y;X_m|S),
\end{equation}
using the definition of (conditional) mutual information. A null player is thus defined as a variable $X_m$ such that $MC_m^v(S) = I(Y;X_m|S) = 0$ for all $S\subseteq V^{-m}$. This definition exactly coincides with the definition of an irrelevant variable (Section~\ref{sec:background:relevance}), since $I(Y;X_m|S)=0$ is equivalent to $Y\indep X_m|S$. Two variables $X_i$ and $X_j$ are symmetric whenever $v(S\cup\{X_i\}) = v(S\cup \{X_j\})$ for all $S\subseteq V^{-i,j}$, which is equivalent to $I(Y;X_i|S) = I(Y;X_j|S)$ for all $S\subseteq V^{-i,j}$, \ie, $X_i$ and $X_j$ bring the same information about $Y$ in all contexts $S$.

With this definition, the following theorem shows that MDI importance of totally randomized trees corresponds to the Shapley value for this TU-game:

\begin{theorem}{(MDI are Shapley values)}\label{THM:MDISHAP}
	For all feature $X_m \in V$, 
	\begin{eqnarray}
	\Imp_{\infty}(X_m) = \phi^{Sh}_v (X_m),
	\end{eqnarray}
	where $\phi^{Sh}_v$ is the Shapley value with $v(S) = I(Y;S)$ ($\forall S\subseteq V$).
\end{theorem}

The proof\footnote{The proofs of all theorems are in Appendix~\ref{app:proofs}.} of this theorem follows from a direct comparison of Equations \ref{eq:shapleyvaldef} and \ref{eqn:popMDI}.

Given this result, the four axioms that uniquely defines Shapley values are obviously satisfied. They translate into the following properties of the importances $\Imp_{\infty}$:

 \textbf{Efficiency}: $\sum_{m=1}^p \Imp_{\infty}(X_m) = v(V) = I(Y;V)$, which states that MDI importances decompose, in an additive way, the mutual information $I(Y;X_1,\dots,X_p)$. This results is identical to Equation~\ref{eq:sumimp}.
  
 \textbf{Symmetry}: If $X_i$ and $X_j$ are symmetric, then $\Imp_{\infty}(X_i)=\Imp_{\infty}(X_j)$. This property is easy to check knowing that $I(Y;S\cup\{X_i\}) = I(Y;S\cup\{X_j\})$ implies that $I(Y;X_i|S)=I(Y;X_j|S)$  and therefore swapping $X_i$ and $X_j$ in Equation~\ref{eqn:popMDI} would keep all terms of the sum unchanged. 

\textbf{Null player}: If $X_m$ is a null player, \ie, an irrelevant variable, then $\Imp_\infty(X_m)=0$. Note that \cite{louppe2013understanding} actually showed a stronger result, stating in addition that $\Imp_{\infty}(X_m)=0$ {\it only if} $X_m$ is irrelevant to $Y$.

\textbf{Strong monotonicity}: Let us assume two outputs $Y_1$ and $Y_2$. Strong monotonicity says that if for all feature subsets $S\subseteq V^{-m}$: $I(Y_1;X_m|S)\ge I(Y_2;X_m|S)$, then we have $\Imp_\infty^{Y_1}(X_m)\ge \Imp_\infty^{Y_2}(X_m)$. This property states that if a variable brings more information about $Y_1$ than about $Y_2$ in all contexts $S$, it is more important to $Y_1$ than to $Y_2$.

The link between MDI importance and Shapley value shows that, in the finite setting, standard MDI importances from totally randomized trees compute an approximation of the Shapley values (for $I(Y;V)$), at least in the categorical setting\footnote{For example, in the context of an ensemble of binary decision trees, MDI importance measure does not estimate the same quantities as in the categorical setting \citep{louppe2014understanding,sutera2019importance}.}. MDI importance will be compared with other measures from the literature that explicitly seek to estimate the same quantities in Section \ref{sec:relatedwork}.

\subsection{Non totally randomized trees}\label{sec:globalmdikgt1}

When $K>1$, $K$ variables are randomly picked at each node and the best split,
in terms of impurity reduction, among these $K$ variables is selected to
actually split the node. Because several variables then compete for each split,
some variables might be never (or less often) selected if there are other
variables providing larger impurity decreases. These masking effects will
impact the properties of the MDI importances. Let us denote by
$\Imp_\infty^K(X_m)$ the importance of $X_m$ derived from randomized trees built
with a given value of $K$ in asymptotic conditions. When $K>1$,
$\Imp_\infty^K(X_m)$ can no longer be decomposed as in
Equation~\ref{eqn:popMDI}, as some $I(Y;X_m|S)$ terms will not be included in
the sum or with a weight different from the one in
Equation~\ref{eqn:popMDI}. Actually, although $\Imp_\infty^K$ attributes a
payoff to each variable in $V$, it can not be interpreted as a TU-value for the
TU-game defined by $(V,v)$, with $v(\cdot) = I(Y;\cdot)$. Indeed, its
computation requires to have access to conditional mutual information of the
form $I(Y;X_m|S=s)$ for all coalition $S$ but also for all set of values $s$ of
variables in $S$ and the latter can not be derived from the knowledge only of
$I(Y;S)$ for all $S$.

The efficiency and null player conditions are however still satisfied by $\Imp_\infty^K$:

\textbf{Efficiency}: \cite{louppe2013understanding} showed that $\sum_{m=1}^p \Imp_\infty^K (X_m) = v(S) = I(Y;V)$ for all $K$ as soon as the trees are fully developed.

\textbf{Null player} If $X_m$ is a null player (\ie, an irrelevant variable), then $\Imp_\infty^K(X_m)=0$  \citep{louppe2013understanding}. Note that $\Imp_\infty^K(X_m)=0$ is however not anymore a sufficient condition for $X_m$ to be irrelevant. It can be shown however that a strongly relevant variable $X_m$ will always be such that $\Imp_\infty^K(X_m)>0$
whatever $K$ \citep{sutera2018random}.

Symmetry is not necessarily satisfied however, since $I(Y;X_i|S)=I(Y;X_j|S)$ for all $S\subseteq V^{-i,j}$ does not ensure that $I(Y;X_i|S=s)=I(Y;X_j|S=s)$ for all $s$, which would be required for the feature to be fully interchangeable when $K>1$. Similarly, strong monotonicity is not satisfied either for $\Imp_\infty^K$, as shown in Example \ref{example:monotonicity} in Appendix~\ref{app:examples}.

Note that as discussed in \citep{louppe2013understanding}, the loss of several
properties when $K>1$ should not preclude using $\Imp_\infty^K$ as an importance
measure in practice. In finite setting, using $K>1$ may still be a sound strategy to guide
the choice of the splitting variables towards the most impactful ones
during tree construction and therefore result in more statistically efficient
estimates.

\section{\textit{Local} MDI importances}\label{sec:local+shapley}

So far, MDI importances are global, in that they assess the overall
importance of each variable independently of any test instance. An important
literature has emerged in the recent year that focuses on local importance
measures that can assess the importance of a variable locally, \ie, for a
specific instance. In Section \ref{sec:localdef}, we define a novel local
MDI-based importance measure. We highlight the main properties of this measure
and show in particular that it very naturally decomposes the standard global
MDI measure. In Section \ref{sec:asymptoticlocal}, we show, in the categorical
setting, that the asymptotic analysis of global MDI can be extended to the
local MDI, which allows us, in Section \ref{sec:localshapley} to show that
local MDI importances are also Shapley values for a specific characteristic
function in the case of totally randomized trees. Finally, in Section
\ref{sec:localandrelevance}, we propose a local adaptation of the notion of
feature relevance and link it with the local MDI measure.

\subsection{Definition and properties}\label{sec:localdef}

Let us denote by $\bmm{x} = (x_1, \dots, x_p)^T$ a given instance of the input variables, with $x_j$ the value of variable $X_j$. In what follows, we will further denote by $\bmm{x}_S$ a given set of values for the variables in a subset $S\subseteq V$ (in particular, $\bmm{x}_{\{X_j\}}=x_j$). 

\begin{definition}{(Local MDI)}\label{def:localmdi}
	The local MDI importance $\Imp(X_m,\bmm{x})$ of a variable $X_m$ with respect to $Y$ for a given instance $\bmm{x}$ is defined as follows 
	\begin{equation}\label{eqn:def:localmdi}
		\Imp(X_m,\bmm{x}) = \dfrac{1}{N_T} \sum_{T}
		\sum_{\substack{t\in T: \nu(s_t)=X_m\\
		\wedge \bmm{x}\in t}}
		i(t) - i(t_{x_m})
	\end{equation} 
	where the outer sum is over the $N_T$ trees of the ensemble, the inner sum is over all nodes that are traversed by $\bmm{x}$ and where $X_m$ is used to split, $t_{x_m}$ is the successor of node $t$ followed by $\bmm{x}$ in the tree (corresponding to $X_m=x_m$), and $i(.)$ is the impurity function.
\end{definition}

This general measure quantifies how important is feature $X_m$ to predict the output of the test example $\bmm{x}$ represented by its input features. It collects all differences $i(t) - i(t_{x_m})$ along all paths traversed by example $\bmm{x}$ in the ensemble. In practice, this can be implemented very efficiently at no additional cost with respect to the computation of a prediction, as soon as all impurities, computed at training time, are stored at tree nodes.

The intuition behind this measure is that a variable is important for a sample $\bmm{x}$ if it leads to important reductions of impurity along the paths traversed by $\bmm{x}$. Note that unlike global MDI, local MDI can be negative, as the impurity can increase from one node to one of its successors. A variable of negative importance for a given sample $\bmm{x}$ is thus such that, in average over all paths traversed by $\bmm{x}$, it actually increases the uncertainty about the output (because it helps for predicting the output of other instances).

A natural link between local and global MDI is given by the following result:
\begin{equation}\label{eq:oilocalmdidecomp}
  \Imp(X_m) = \frac{1}{N} \sum_{i=1}^N \Imp(X_m,\bmm{x}^i),
\end{equation}
where $\{(\bmm{x}^1,y^1),\dots,(\bmm{x}^N,y^N)\}$ is the learning sample of $N$ examples that was used to grow the ensemble of trees. This result can be shown easily by combining the definitions in Equations \ref{eqn:mdi_tree} and \ref{eqn:def:localmdi} of global and local MDI respectively. Local MDI is thus a way to decompose the global MDI over all training examples.

\subsection{Asymptotic analysis}\label{sec:asymptoticlocal}

In the categorical setting and asymptotic conditions, the decomposition in Equation~\ref{eqn:popMDI} for totally randomized trees can be adapted to the local MDI measure, denoted $\Imp_\infty(X_m,\bmm{x})$.

\begin{theorem}{(Asymptotic local MDI)} \label{THM:LOCAL} The local MDI importance $\Imp_\infty(X_m,\bmm{x})$ of a variable $X_m$ with respect to $Y$ for a given sample $\bmm{x}$ as computed with an infinite ensemble of fully developed totally randomized trees and an infinitely large training sample is 
	\begin{multline}
	\label{THM:LOCAL:EQN}
	\small
		\Imp_\infty(X_m,\bmm{x}) = \sum_{k=0}^{p-1} \dfrac{1}{C^k_p} \dfrac{1}{p-k} \sum_{B\in \mathcal{P}(V^{-m})} H(Y|B=\bmm{x}_B)- H(Y|B=\bmm{x}_B, X_m= x_m)
	\end{multline}
\end{theorem}
{where $H(Y|\cdot)$ is the conditional entropy of $Y$.}
Similarly as in the finite setting, $\Imp_\infty(X_m,\bmm{x})$ can be negative,
since the difference $H(Y|B=\bmm{x}_B)-H(Y|B=\bmm{x}_B,X_m=x_m)$ is not always
positive. Example~\ref{example:neg} in Appendix~\ref{app:examples} illustrates one such a situation.

In asymptotic condition, the decomposition in Equation~\ref{eq:oilocalmdidecomp} furthermore becomes:
\begin{equation}
  \Imp_\infty(X_m) = \sum_{\bmm{x}\in {\cal V}} P(V=\bmm{x}) \Imp_\infty(X_m,\bmm{x}),
\end{equation}
where the sum is over all possible input combinations. Combined with \ref{eq:sumimp}, this leads to the following double decomposition (over features and instances) of the information $I(V;Y)$:
\begin{equation}\label{eq:localdecompinfo}
I(V;Y)  = \sum_{m=1}^p\sum_{\bmm{x}\in\mathcal{V}} P(V=\bmm{x}) \Imp_\infty(X_m,\bmm{x}).
\end{equation}

\subsection{Local MDI importances are Shapley values}\label{sec:localshapley}

Let us define a \textit{local} characteristic function $v_{loc}(S;\bmm{x}) = H(Y)-H(Y|S=\bmm{x}_S)$, which measures the decrease in uncertainty (\ie, the amount of information) about the output $Y$ when the variables in $S$ are known to be $\bmm{x}_S$. This characteristic function is thus parameterized by $\bmm{x}$. The proof of Theorem~\ref{THM:MDISHAP} can be adapted to the decomposition in Equation~\ref{THM:LOCAL:EQN} to show that local MDI importances of totally randomized trees in asymptotic conditions are Shapley values with respect to $v_{loc}(.;\bmm{x})$:
\begin{equation}
\Imp_\infty(X_m,\bmm{x}) = \phi^{Sh}_{v_{loc}(.;\bmm{x})}(X_m)
\end{equation}
As a consequence, $\Imp_\infty(X_m,\bmm{x})$ satisfies the Shapley value properties at any point $\bmm{x}$, \ie:

 \textbf{Efficiency}: $v_{loc}(V;\bmm{x})=H(Y)-H(Y|V=\bmm{x}) = \sum_{m=1}^p \Imp_\infty(X_m,\bmm{x})$. This is in accordance with the decomposition in Equation~\ref{eq:localdecompinfo}, since $\sum_{\bmm{x}\in{\cal V}} P(V=\bmm{x}) v_{loc}(V;\bmm{x}) = \sum_{x\in {\cal V}} P(V=\bmm{x}) (H(Y) - H(Y|V=\bmm{x})) = I(V;Y)$.

\textbf{Symmetry}: If $X_i$ and $X_j$ are such that $H(Y)-H(Y|S=\bmm{x}_S,X_i=x_i) = H(Y)-H(Y|S=\bmm{x}_S,X_j=x_j)$ for every $S\subseteq V^{-i,j}$, then $\Imp_\infty(X_i,\bmm{x})=\Imp_\infty(X_j,\bmm{x})$.

\textbf{Null player} If $X_i$ is such that $H(Y|S=\bmm{x}_S,X_m=x_m)=H(Y|S=\bmm{x}_S)$ for all $S\subseteq V^{-m}$, then $\Imp_\infty(X_i,\bmm{x})=0$.

\textbf{Strong monotonicity}: Assuming two outputs $Y_1$ and $Y_2$, if $H(Y_1|S=\bmm{x}_S)-H(Y_1|S=\bmm{x}_S,X_m=x_m)\ge H(Y_2|S=\bmm{x}_S)-H(Y_2|S=\bmm{x}_S,X_m= x_m)$ for all $S\subseteq V^{-m}$, then we have $\Imp^{Y_1}(X_m,\bmm{x})\ge \Imp^{Y_2}(X_m,\bmm{x})$. 

As in the case of global MDI, using non totally randomized trees ($K>1$) will make local MDI to depart from the Shapley values, because of masking effects. Actually, local MDI importances will again not correspond to TU-values for $v_{loc}(\cdot;\bmm{x})$, since they are not uniquely defined by $v_{loc}(\cdot;\bmm{x})$. Indeed, tree splits along the paths traversed by $\bmm{x}$ can not be determined from $v_{loc}(\cdot;\bmm{x})$ only, as they depend on impurity reductions on other paths as well. However, the efficiency and null player properties will again remain valid, although symmetry and strong monotonicity are not guaranteed.

\subsection{Local relevance}\label{sec:localandrelevance}

A major result regarding the global MDI importance in asymptotic conditions is its link with \cite{kohavi1997wrappers}'s notion of feature relevance. A similar relationship can be established between local MDI importance measures and a new notion of local relevance (at the level of a samples) inspired from the null player property of Shapley values.

\begin{definition}
\label{def:localirrelevance}
    $X_m$ is \textbf{locally irrelevant} at $\bmm{x}$ with respect to the output $Y$ iff 
    $P(Y=y|X=x_m,B=\bmm{x}_B)=P(Y=y|B=\bmm{x}_B)$
for all $B\subseteq V^{-m}$ and all $y\in {\cal Y}$. It is \textbf{locally relevant} otherwise. 
\end{definition}
A variable is thus locally irrelevant at $\bmm{x}$ if knowing its values never changes the probability of any output whatever the other variables that are known. Local relevance can be linked with global relevance through the following theorem.

\begin{theorem}
\label{THM:IRR}
A variable $X_m$ is irrelevant with respect to $Y$ if and only if it is locally irrelevant with respect to $Y$ for all $\bmm{x}$ such that $P(V=\bmm{x})>0$.
\end{theorem}

In the categorical setting and asymptotic conditions, local relevance is linked to local MDI through the following theorem:
\begin{theorem}\label{COR:=0}
  If a variable is \textit{locally irrelevant} at $\bmm{x}$ with respect to $Y$, then $\Imp_\infty(X_m,\bmm{x}) = 0$.
\end{theorem}
Theorem 4 coincides exactly with the null player property of Section \ref{sec:localshapley}. Local MDI importance is thus a sensible score to identify locally irrelevant variables. Note that, unlike with global MDI, there might exist variables $X_m$ such that $\Imp_{\infty}(X_m, \bmm{x})=0$ despite $X_m$ being locally relevant at $\bmm{x}$. However, a globally relevant variable $X_m$ will always receive a non zero $\Imp_{\infty}(X_m, \bmm{x})$ at some $\bmm{x}$. 

\section{Discussion and related works}\label{sec:relatedwork}

In the literature, Shapley values have been mostly used to decompose model predictions $\hat{f}(\bmm{x})$ at any $\bmm{x}$ into a sum of terms that represent the (local) contribution of each variable to the prediction \citep{strumbelj2010,lundberg2017unified}. The characteristic function $v_{\hat{f}}$ considered by these methods is $v(S) = \hat{f}_S(\bmm{x}_S) - \hat{f}_\emptyset(\bmm{x}_\emptyset)$, where $\hat{f}_S(\bmm{x}_S)$ is the model to be explained restricted to the variables in $S$ and $\hat{f}_\emptyset(\bmm{x}_\emptyset)$ is often set to $E[\hat{f}(X)]$.  Typically, $\hat{f}_S(\bmm{x}_S)$ is defined as $\mathbb{E}[\hat{f}(X)|X_S=\bmm{x}_S]$, where the expectation is over the conditional $p(X_{\bar{S}}|X_S=\bmm{x}_S)$ or the marginal (a.k.a. interventional) $p(X_{\bar{S}})$ distribution. These methods are mostly model agnostic, \ie, they can handle any machine learning model, considered as a black-box, although the estimation of restricted models and the computation of the Shapley values (Equation \ref{eq:shapleyvaldef}) can be very challenging in general.

Among this literature, \cite{lundberg2020local} have proposed TreeSHAP, a framework to efficiently compute Shapley values when $\hat{f}$ are trees or sum of trees. 
One of the only alternative local importance measures for trees is Saabas' heuristic method (implemented in \citep{saabas2014interpreting}). Saabas' method measures local variable importances for a sample $\bmm{x}$ by collecting the changes in the (expected) model prediction due to each variable value along the tree branches followed by $\bmm{x}$. Like TreeSHAP, Saabas' importances sum to the model prediction at $\bmm{x}$. They are much faster to compute but do not satisfy all properties of Shapley values, in particular strong monotonicity.

One main difference between local MDI and TreeSHAP/Saabas as studied in \citet{lundberg2020local} is that local MDI decomposes entropy (or more generally impurity) reductions ($v_{loc}(V;\bmm{x})$), while TreeSHAP/Saabas decompose model predictions\footnote{{Although TreeSHAP authors advocate the decomposition of model predictions as the way to go, a variant of TreeSHAP \citep{lundberg2020local} can also decompose model loss by enforcing feature independence, at a higher computational cost however than the local MDI measure proposed here.}} ($v_{\hat{f}}$). As a consequence, local MDI scores are independent of output normalisation or scaling and do not require to choose a specific class probability score to be used as $\hat{f}(\bmm{x})$ to be decomposed. This also allows to connect local and global MDI in a natural way, and gives a probabilistic interpretation to the null player property in terms of variable irrelevance. Algorithmically, local MDI uses the exact same collecting procedure along tree paths as Saabas' measure, replacing output differences with impurity reductions. Similarly as Saabas, local MDI results in a much more efficient and simpler estimation scheme than TreeSHAP but it looses some properties of Shapley values in the general case. Section \ref{sec:localshapley} however shows that these properties are retrieved in the case of totally randomized trees, at least asymptotically. This also applies to Saabas' measure (as sketched in Theorem 1 in the supplement of \citet{lundberg2020local}). Although TreeSHAP is guaranteed to ensure strong monotonicity asymptotically, the relevance of its scores is still tied to the quality of the tree-based model that it explains. For example, using TreeSHAP with Random forests with $K>1$ or pruned trees will also potentially lead to biases in the importance scores (\eg, due to masking effects) with respect to what would be obtained if $\hat{f}$ was the Bayes classifier. We believe this is a similar trade-off as the one met in local MDI with respect to $K$.

Our results also highlight a link between global MDI and SAGE \citep{covert2020understanding}, a purely model-agnostic method for global importance computation. SAGE estimates Shapley values for $v_{\ell}(S) = \mathbb{E}[\ell(\hat{f}_\emptyset(X_\emptyset),Y)] - \mathbb{E}[\ell(\hat{f}_S(X_S),Y)]$ where $\ell$ is a loss function and expectations are taken over $p(V,X)$. \cite{covert2020understanding} have shown that when $\ell$ is cross-entropy, $\hat{f}$ is the Bayes classifier, and restricted models are estimated through the conditional distributions, then the population version of SAGE is strictly identical to Equation \ref{eqn:popMDI}. Interestingly, both methods arrive to this population formulation through very different algorithms. SAGE explicitly estimates Shapley values, while global MDI are obtained by collecting impurity reductions at tree nodes in a random forest. Global MDI departs from \ref{eqn:popMDI} and Shapley values when $K>1$. On the other hand, being model agnostic, like TreeSHAP, SAGE is tied to the quality of the model it explains. Given the difficulty of sampling from the conditional distribution, its implementation also samples from the  marginal distribution instead, which makes it depart from Shapley values and affects its convergence to \ref{eqn:popMDI}. In practice, we will show in the next section that both methods produce very similar results, when used with Random forests (but with a strong advantage to global MDI in terms of computing times).

Overall, we believe our analysis of local and global MDI sheds some new lights on these measures. Although they were not designed as such, these methods can indeed be interpreted as procedures to sample variable subsets and compute mutual information such that they provide estimates of Shapley values for a very natural characteristic function based on mutual information. Although they are tightly linked algorithmically with Random forests, they actually highlight general properties of the original data distribution independently of these models. This makes them very different from model-agnostic methods that explain pre-existing models, furthermore regardless of the data distribution when restricted models are estimated from marginal distributions.

\section{Illustrations}\label{sec:illustration}

Here, global and local MDI are illustrated and compared against SAGE, Saabas, and TreeSHAP on two classification problems. The \texttt{led} problem \citep{breiman1984classification} consists of an output $Y\in \{0,\ldots, 9\}$ with equal probability and seven binary variables representing each a segment of a seven-segment display and whose values are determined unequivocally from $Y$. The \texttt{digits} problem consists of an output $Y\in \{0,\ldots, 9\}$ and 64 integer inputs corresponding to the pixels of a $8\times 8$ grayscale image of a digit. Additional experiments on other datasets are reported in Appendix~\ref{app:supplementaryresults_local}. In all experiments, importance scores are computed either using Scikit-Learn \citep{pedregosa2011scikit} or method authors' original code (data and code are open-source, see details in Appendix~\ref{app:code}).

\begin{figure*}
\begin{minipage}{.58\textwidth}

    \centering
    \hfill

    \includegraphics[width=0.49\linewidth]{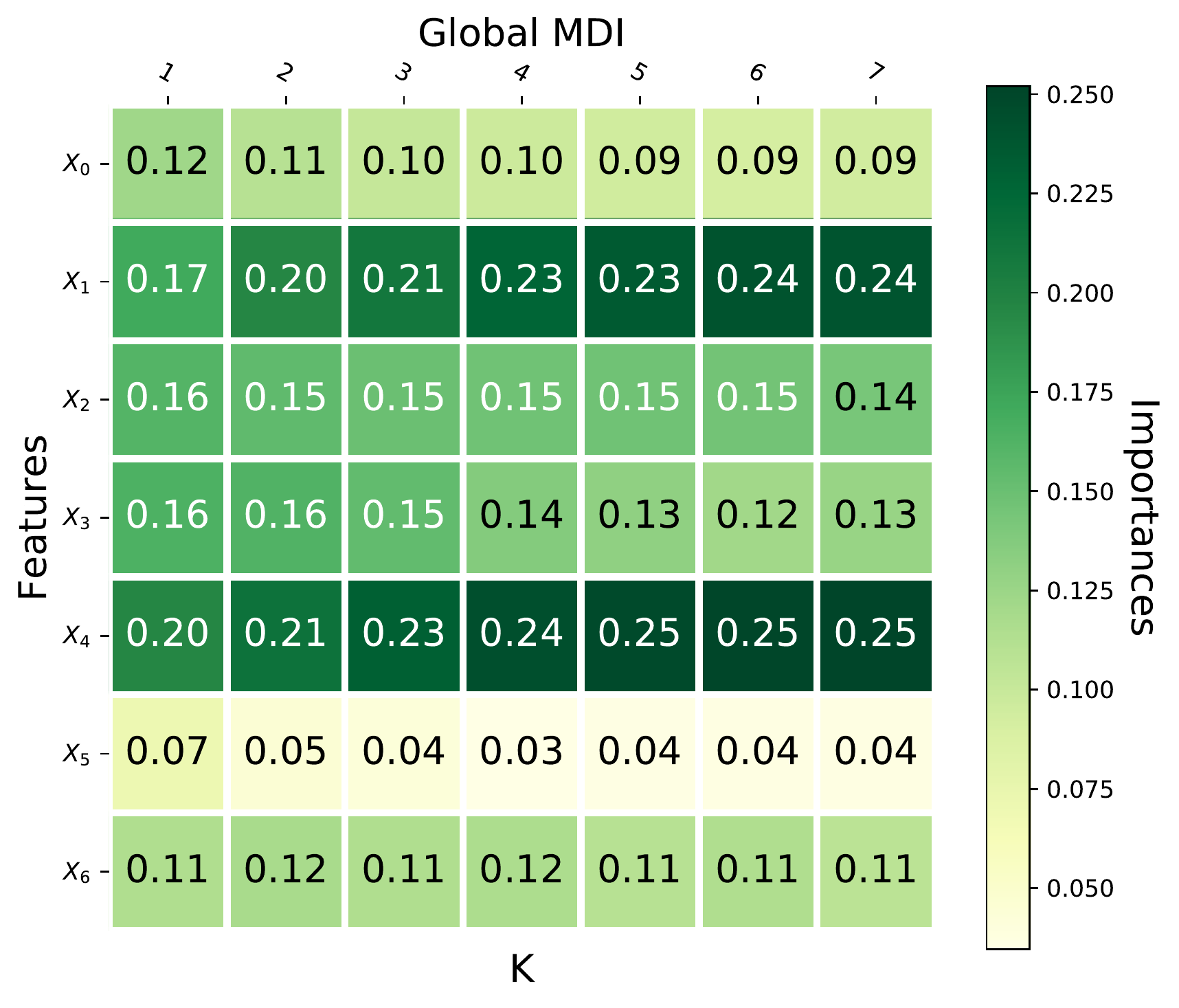}
    \includegraphics[width=0.49\linewidth]{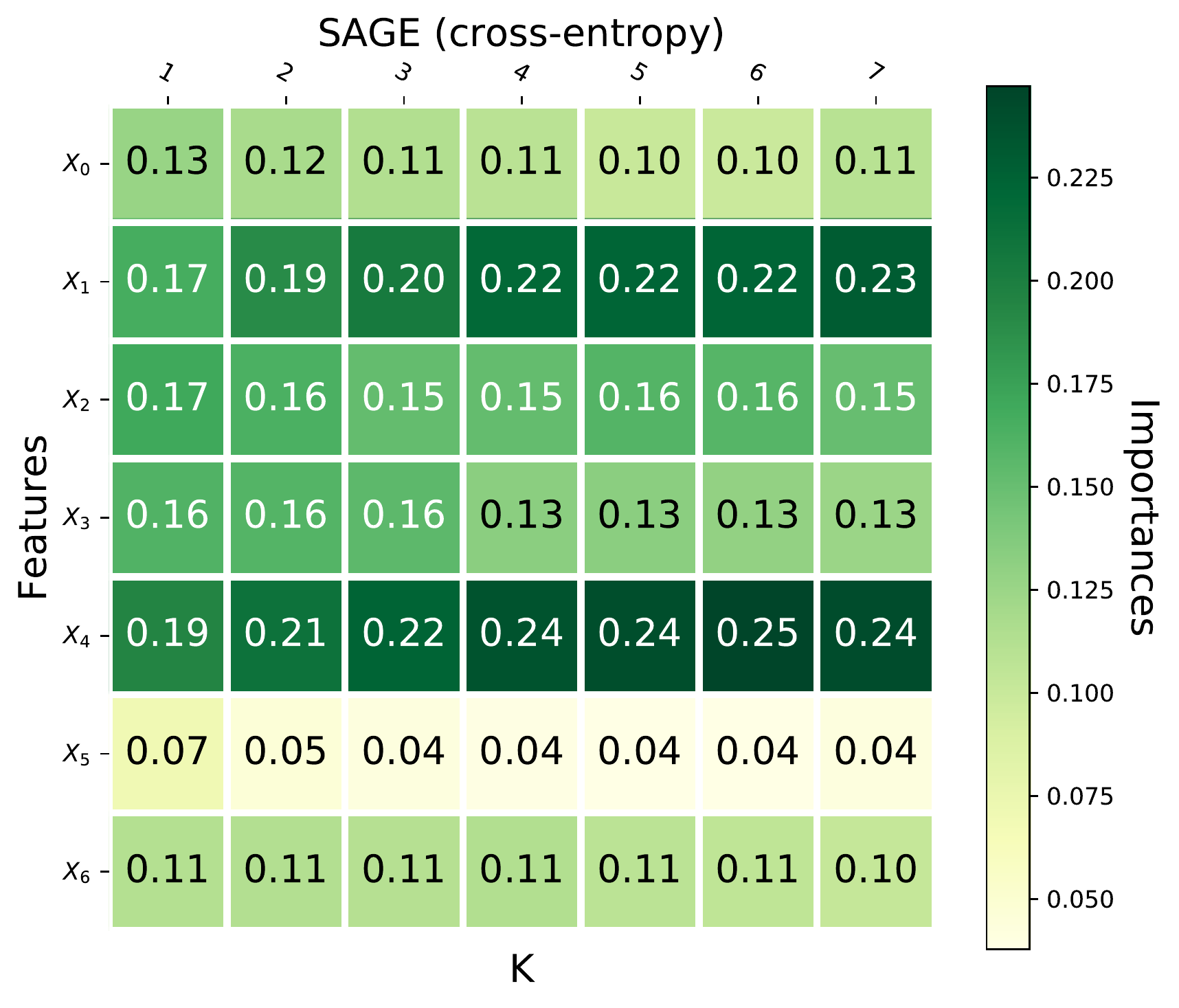}
    \caption{Normalized importance scores derived from an ensemble of totally and non-totally randomized Extra-Trees (with $K=1,\dots,p$) for the global MDI importance measure (left) and SAGE (right).}\hfill
    \label{fig:Kp:MDI+SAGE}

\end{minipage}\hfill
\begin{minipage}{.38\textwidth}

    \centering
    \includegraphics[width=0.73\linewidth]{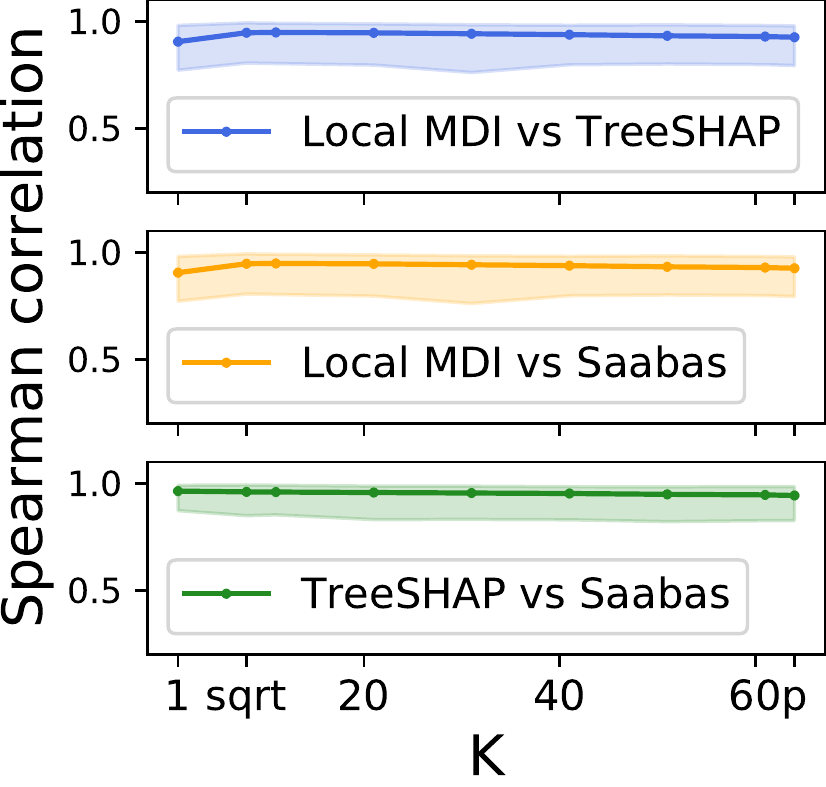}
    \caption{Mean correlation (over all samples)  w.r.t. increasing $K$ between absolute importance scores for \texttt{digits} ([min,max] is shaded). }
    \label{fig:local:corr+K}

\end{minipage}
\vspace{-0.2cm}
\end{figure*}
\begin{figure}[tbp]
\centering
\includegraphics[width=0.48\linewidth]{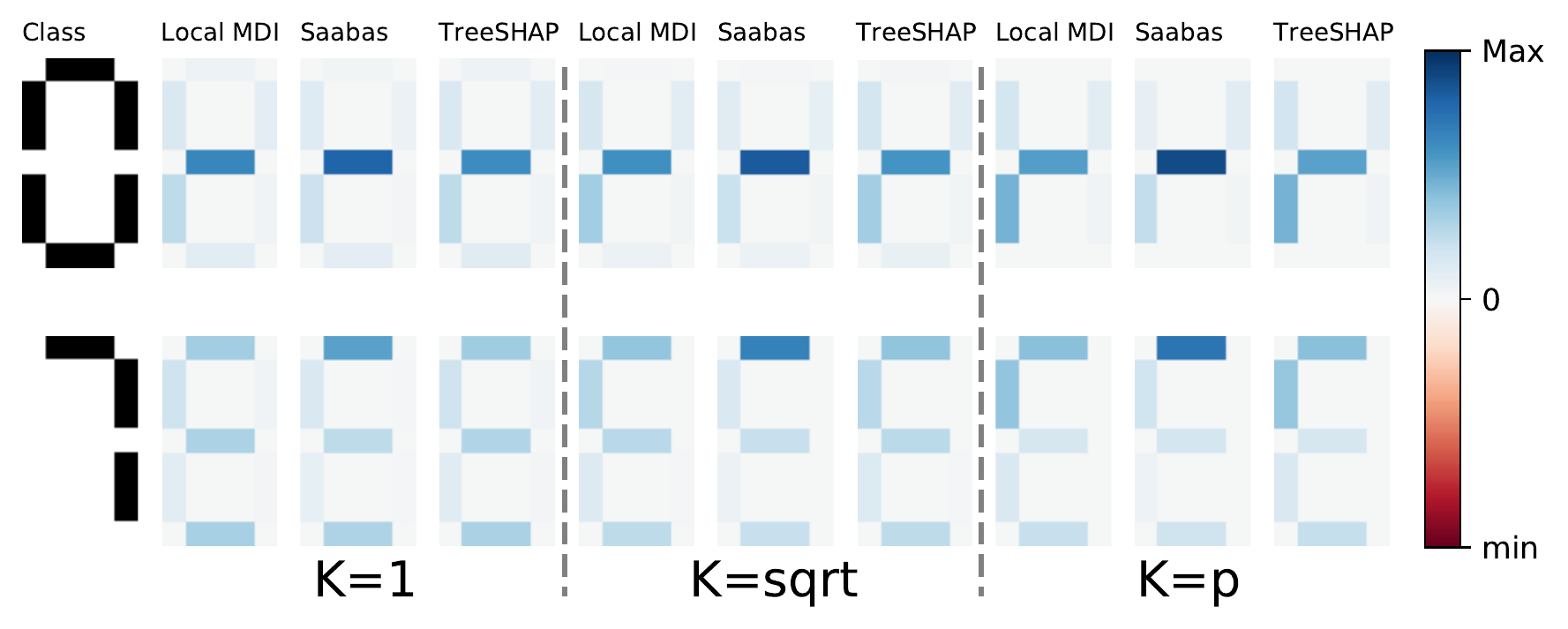}
\includegraphics[width=0.495\linewidth]{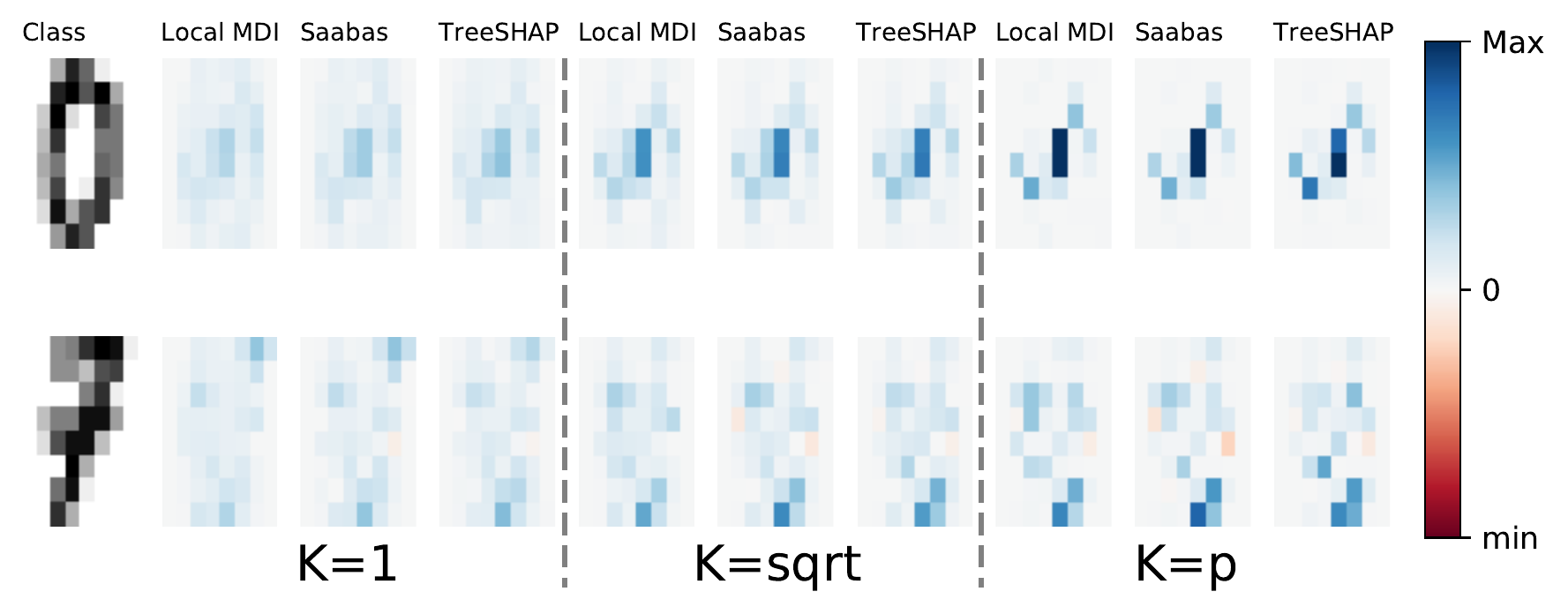}
\caption{Local importances derived by local measures from a forest of $1000$ Extra-Trees with $K\in\{1,\sqrt{p},p\}$) for \texttt{led} (left) and \texttt{digits} (right). Results for all samples can be found in Appendix~\ref{app:supplementaryresults_local}.}
\label{fig:exp:local}
\end{figure}

{\bf Global importances} \label{sec:illustration:global}
are shown in Figure \ref{fig:Kp:MDI+SAGE} on the \texttt{led} dataset for an ensemble of 1000 Extra-Trees \citep{geurts2006extremely} and several values of $K$, with global MDI and SAGE (using the cross-entropy loss). Importances are normalized such that the sum of (absolute values of) the scores is equal to one. With $K=1$, both approaches clearly yield very similar scores, as expected from the discussion in Section \ref{sec:relatedwork}. Both methods remain very close when $K$ is increased. Regardless of the importance measure, one can notice the masking effect that favors $X_1$ and $X_4$ at the expense of the other variables when $K$ increases. The same effect appears when SAGE uses another loss function and for the global importance measure derived from TreeSHAP (see Appendix~\ref{app:supplementaryresults_global}).

{\bf Local importances} are then computed for both classification problems from an ensemble of 1000 Extra-Trees and with the three local importance measures. Scores from Saabas and TreeSHAP are obtained from the additive decomposition of the conditional probability of the predicted class (which might not be the true class). Local feature importances for two samples are reported in Figure~\ref{fig:exp:local} by coloring either the corresponding segment (left) or pixel (right) for three values of $K$. The three local importance measures yield very similar importance scores, suggesting that they provide matching explanations of model predictions. This similarity is further quantified by measuring the correlation between the (absolute value of) local importance scores of all pairs of methods for several $K$ (Figure~\ref{fig:local:corr+K} for \texttt{digits}, Appendix~\ref{app:supplementaryresults_local} for \texttt{led}). The mean correlation over all samples remains close to 1 for all pairs, although the variation (depicted by the shaded area delimiting the range between minimal and maximal correlations) is impacted by the value of $K$. As expected, Saabas and TreeSHAP, which both decompose model predictions, are closer to each other than to Local MDI.

\section{Conclusion} \label{sec:conclu}
MDI importances have  been  used extensively as a way of measuring, globally, the respective contribution of variables. In this paper, we showed that global MDI importances derived from totally randomized trees are actually Shapley values that decompose the mutual information $I(V;Y)$ when the impurity measure is the Shannon entropy. We then proposed a local MDI importance that very naturally decomposes global MDI over the training examples. We showed that local MDI importances are also Shapley values with respect to conditional entropy reductions and that they are consistent with respect to a novel local relevance notion. We compared MDI importances conceptually with other recent local and global feature importance scores inspired from Shapley values and showed empirically that all these methods are very close, while both global and local MDI importances do not require any extra computation with respect to the tree construction. Overall, local and global MDI measures provide a natural and efficient way of explaining properties of the data distribution.

While the main results of this paper assume categorical variables and Shannon entropy as impurity measure, they can be extended to other impurity measures and to regression (see Appendix \ref{app:generalisation}). As future works, we would like to better characterize these measures in non-asymptotic conditions and outside of the purely categorical setting. More experiments should be also carried out to better highlight differences, practically, between the MDI family and other methods such as TreeSHAP and SAGE that more explicitly approximate Shapley values. Finally, the link with Shapley values and TU-games in general could be further investigated to propose other extensions of MDI measures (for example to highlight variables interactions as in \citep{lundberg2020local}).

\subsection*{Acknowledgments}
 Antonio Sutera is supported via the Energy Transition Funds project EPOC 2030-2050 organized by the FPS economy, S.M.E.s, Self-employed and Energy. This work was partially supported by Service Public de Wallonie Recherche under grant n$^{\circ}$ 2010235 – ARIAC by DIGITALWALLONIA4.AI.
 
\bibliographystyle{plainnat}
\bibliography{reference}

\newpage
\appendix
\section{Proofs}\label{app:proofs}
\subsection{Proof of Theorem \ref{THM:MDISHAP}}
\textbf{Theorem 1.}
\textit{{(MDI are Shapley values)}
	For all feature $X_m \in V$, 
	\begin{equation}
	Imp_{\infty}(X_m) = \phi^{Sh}_v (X_m), \tag{8}
	\end{equation}
	where $\phi^{Sh}_v(X_m)$ is the Shapley value of $X_m$ with respect to the characteristic function $v(S) = I(Y;S)$ (with $S\subseteq V$).}

\begin{proof}
	Let us first note that 
	\begin{eqnarray}
	v(S\cup\{X_m\}) - v(S) &=& I(Y;S,X_m) - I(Y;S)\notag\\
	&=& H(Y) - H(Y|S,X_m) - H(Y) + H(Y|S)\notag\\
	&=& I(X_m;Y|S) \label{eqn:proof:ThmMDIShap:1}
	\end{eqnarray}
	Replacing the characteristic functions as defined in this equation in Equation \ref{eq:shapleyvaldef} of the main paper, Shapley values can thus be defined as: 
	\begin{eqnarray}
	\phi_v(X_m) = \sum_{S\subseteq V^{-m}} \dfrac{|S|!(p-|S|-1)!}{p!} I(X_m;Y|S)
	\end{eqnarray}
	The sum can be reorganized according to the size of the subsets $S$ from $V^{-m}$:
	\begin{eqnarray}
	\phi_v(X_m) = \sum_{k=0}^{p-1} \dfrac{k!(p-k-1)!}{p!} \sum_{S\subseteq \mathcal{P}_k(V^{-m})} I(X_m;Y|S)
	\end{eqnarray}
	which is strictly equivalent to $Imp_\infty(X_m)$ given that:
	\begin{eqnarray}
	\dfrac{k!(p-k-1)!}{p!} &=& \dfrac{(p-k)!k!}{p!}\dfrac{1}{p-k}\notag\\
	&=&\dfrac{1}{\dfrac{p!}{(p-k)!k!}}\dfrac{1}{p-k}\notag\\
	&=&\dfrac{1}{C^k_p} \dfrac{1}{p-k}
	\end{eqnarray}
\end{proof}

\subsection{Proof of Theorem \ref{THM:LOCAL}}
\textbf{Theorem 2.}
\textit{
{(Asymptotic local MDI)} The local MDI importance $Imp_\infty(X_m,\bmm{x})$ of a variable $X_m$ with respect to $Y$ for a given sample $\bmm{x}$ as computed with an infinite ensemble of fully developed totally randomized trees and an infinitely large training sample is 
	\begin{multline}
	\tag{11}
	\small
		Imp_\infty(X_m,\bmm{x}) = \sum_{k=0}^{p-1} \dfrac{1}{C^k_p} \dfrac{1}{p-k} \sum_{B\in \mathcal{P}(V^{-m})} H(Y|B=\bmm{x}_B)- H(Y|B=\bmm{x}_B, X_m= x_m)
	\end{multline}
}
\begin{proof}
	Let $B(t)=(X_{i_1},\dots,X_{i_k})$ be the subset of $k$ variables tested in the branch from the root node to the parent of $t$ and $b(t)$ be the vector of values of these variables. As the number of training samples grows to infinity, the probability that a sample reaches node $t$ is $P(B(t)=b(t))$ (according to  $P(X_1,\dots,X_p,Y)$). As the number $N_T$ of totally randomized trees also grows to infinity, the local importance of variable $X_m$ for sample $\bmm{x}^i$ can then be written:
	\begin{eqnarray}\label{eqn:thm:asyloc:proof1}
	   Imp(X_m,\bmm{x}) = \sum_{B\subseteq V^{-m}} \beta\left ( H(Y|B=\bmm{x}_b) - H(Y|B=\bmm{x}_B,X_m=x_{m}) \right )  
	\end{eqnarray}
	where $\beta$ is probability that a node $t$ (at depth $k$) in a totally randomized tree tests the variable $X_m$ and is such that $B(t)=B$ and $b(t)=\bmm{x}_B$. $\beta$ is given by \citep{louppe2013understanding} as being equal to $\frac{1}{C_p^k}\frac{1}{p-k}$ and remains valid because it only depends on the size $k$ of $B$ and on the number $p$ of variables. Notice already the similarity with the intermediate formulation in the proof of Theorem~1 from \citep{louppe2013understanding} where Equation~\ref{eqn:thm:asyloc:proof1} reduces the inner sum to a single term, the one corresponding to the given $b=\bmm{x}_B$. Rewritting Equation~\ref{eqn:thm:asyloc:proof1} in order to group subsets $B$ according to their sizes, we have
    \begin{eqnarray}\small
		Imp_\infty(X_m,\bmm{x}) = \sum_{k=0}^{p-1} \dfrac{1}{C^k_p} \dfrac{1}{p-k} \sum_{B\in \mathcal{P}(V^{-m})} H(Y|B=\bmm{x}_B)   -H(Y|B=\bmm{x}_B, X_m=\bmm{x}_{m})
	\end{eqnarray}
    where the inner sum is over the set of subsets of $V^{-m}$ of cardinality $k$ (\ie, the different paths of length $k$ leading to a test on $X_m$), and 
    which completes the proof.
\end{proof}

\subsection{Proof of Theorem \ref{THM:IRR}}

\textbf{Theorem 3.}\textit{
{(Equivalence of irrelevance)}
A variable $X_m$ is irrelevant with respect to $Y$ if and only if it is locally irrelevant with respect to $Y$ for all $\bmm{x}$ such that $P(V=\bmm{x})>0$.
}

\begin{proof}
{This proof directly stems from the following intuitive observation: the irrelevance property considers all $\boldsymbol{x}$ while the local irrelevance one only considers one $\boldsymbol{x}$. 
If local irrelevance is satisfied for all $\boldsymbol{x}$, then irrelevance is satisfied. The other way around is trivial. Thus the irrelevance property is equivalent to the set of local irrelevance properties corresponding to each $\boldsymbol{x}$. Mathematically, we can also prove this equivalence as follows.}
By definition of irrelevance, $X_m\indep Y|B$ for all $B\subseteq V^{-m}$, where $X_m\indep Y|B$ is the conditional independence and is equivalent (in the case of discrete variables and assuming $P(B=b)>0$) to saying that $P(X_m=x_m,Y=y|B=b)=P(X_m=x_m|B=b)P(Y=y|B=b)$ for all $y,x_m$. It comes that $P(Y=y|X_m=x_m,B=b)=P(Y=y|B=b)$ and so both notions are equivalent if the local definition is valid for all $y$ and $x_m$. 

Note that this proof can be extended to continuous variables by changing probabilities $P(X=x)$ to $P(X \leq x)$.
\end{proof}

\subsection{Proof of Theorem \ref{COR:=0}}

\textbf{Theorem 4.}
\textit{
  If a variable is \textit{locally irrelevant} at $\bmm{x}$ with respect to $Y$, then $Imp_\infty(X_m,\bmm{x}) = 0$.
}
\begin{proof}
The proof stems from the definition of the local irrelevance. By definition, if $X_m$ is locally irrelevant at $\bmm{x}$ with respect to the output $Y$, then $P(Y=y|X_m=x_m,B=\bmm{x}_B)=P(Y=y|B=\bmm{x}_B)$ for all $B\subseteq V^{-m}$ and all $y\in\mathcal{Y}$. Consequently, 
{\begin{eqnarray*}
  H(Y|B=\bmm{x}_B,X_m=x_m)&=&-\sum_{y\in\mathcal{Y}} P(y|\bmm{x}_B,x_m)\log{P(y|\bmm{x}_B,x_m)}\\
  &=&-\sum_{y\in\mathcal{Y}} P(y|\bmm{x}_B)\log{P(y|\bmm{x}_B)}\\
  &=&H(Y|B=\bmm{x}_B).
\end{eqnarray*}}

If $X_m$ is locally irrelevant at $\bmm{x}$, each collected term is therefore equal to zero, leading to $Imp_\infty(X_m,\bmm{x})=0$.

\end{proof}

\section{Examples}\label{app:examples}
\begin{example}\label{example:monotonicity} Let us consider the two binary classification problems, with outputs $Y_1$ and $Y_2$ and two binary inputs $X_1$ and $X_2$, described in Table \ref{tab:exmonotonicity}. One can compute the following conditional mutual information terms:
  \begin{eqnarray*}
    I(Y_1;X_1) =  0.091 & (\geq) & I(Y_2;X_1) = 0.002,\\
    I(Y_1;X_1|X_2) = 0.269 & (\geq) & I(Y_2;X_1|X_2) = 0.243,\\
    I(Y_1;X_2) = 0.002 & (\leq) &  I(Y_2;X_2) = 0.016, \\
    I(Y_1;X_2|X_1) = 0.180 & (\leq) & I(Y_2;X_2|X_1) = 0.258.\\
  \end{eqnarray*}
  If $K=2$, the forest reduces to a single tree both for $Y_1$ and $Y_2$. For $Y_1$, this tree first splits on $X_1$ and then on $X_2$, resulting in the following importances:
  \begin{eqnarray*}
    Imp^{K=2,Y_1}_\infty(X_1) & = & I(Y_1;X_1) = 0.091,\\
    Imp^{K=2,Y_1}_\infty(X_2) & = & I(Y_1;X_2|X_1) = 0.180.
  \end{eqnarray*}
  For $Y_2$, the tree first splits on $X_2$ and then on $X_1$, resulting in the following importances:
  \begin{eqnarray*}
    Imp^{K=2,Y_2}_\infty(X_1) & = & I(Y_2;X_1|X_2) = 0.243,\\
    Imp^{K=2,Y_2}_\infty(X_2) & = & I(Y_2;X_2) = 0.016.
  \end{eqnarray*}
  One can see that the strong monotonicity property is violated both
  for $X_1$ and $X_2$. For example, although $X_1$ brings more
  information about $Y_1$ than about $Y_2$ in all contexts, it is more
  important to $Y_2$ than $Y_1$. This is due to the fact that in the
  tree for $Y_2$, $X_1$ appears at the second level of the tree and it
  thus receive more credit than in the tree for $Y_1$ where it appears
  at the top node.
  
  \begin{table}[h!]
    \caption{Definition of two outputs for which the strong monotonicity constraint is not satisfied, neither for $X_1$, nor for $X_2$. All input combinations are assumed to be equiprobable.\label{tab:exmonotonicity}}
\begin{center}
    \begin{tabular}{cc|cc}
      $X_1$& $X_2$& $P(Y_1=1|X_1,X_2)$& $P(Y_2=1|X_1,X_2)$\\
      \hline
0  &0  &0.1 &0.1\\
0  &1  &0.5 &0.8\\
1  &0  &0.9 &0.7\\
1  &1  &0.4 &0.3
    \end{tabular}
\end{center}
  \end{table}
\end{example}

\begin{example}\label{example:neg}
  Let us consider a binary classification problem with a single binary input $X_1$ and assume that $P(Y=0)>P(Y=1)$ and $P(Y=0|X_1=0)=P(Y=1|X_1=0)$. In this case, $Imp_\infty(X_1,0) = H(Y) - H(Y|X_1=0)$, which is negative. Table~\ref{tab:exneg} gives a numerical example of this.
\end{example}
  \begin{table}[h!]
    \caption{Definition of $X_1$ and $Y$ such that $Imp_\infty(X_1,0)<0$. Here, $Imp_\infty(X_1,0)=0.81-1=-0.19$. All input combinations are assumed to be equiprobable.\label{tab:exneg}}
\begin{center}
    \begin{tabular}{c|cc|c}
      $X_1$& $P(Y=0|X_1)$& $P(Y=1|X_1)$&\\
      \hline
0  &0.5 & 0.5 & $H(Y|X_1=0)= 1$\\
1  &1.0 & 0.0 & $H(Y|X_1=1)= 0$
    \end{tabular}\hfill
    \begin{tabular}{c|c|c}
      & $P(Y)$&\\\hline
    0  &$0.75$ &\multirow{2}{*}{$H(Y)=0.81$}\\
    1  &$0.25$ &
    \end{tabular}
\end{center}
  \end{table}
  
\section{Code and data}\label{app:code}
Tree-based models are computed using Scikit-Learn \citep{pedregosa2011scikit} (BSD-3-Clause License) and other importance measures are computed using the corresponding latest available source code: Saabas (BSD-3-Clause License) from \textsc{treeinterpreter} \textit{`v0.1.0'}, TreeSHAP from \textsc{shap} `v0.38.2' (MIT License), and SAGE from its Github repository (MIT License). As there is no versioning of the \textsc{sage} package, the code used was lastly downloaded on the 6$^{th}$ of February 2021. Global MDI is computed using Scikit-Learn. Source code (BSD-3-Clause License) to compute local MDI is available (open-source) at \url{https://github.com/asutera/Local-MDI-importance}.

\section{Supplementary results for global importance measures}\label{app:supplementaryresults_global}

\begin{figure*}[ht]
    \centering
    \hfill
    \subfloat[SAGE (mse)]{\includegraphics[width=0.45\linewidth]{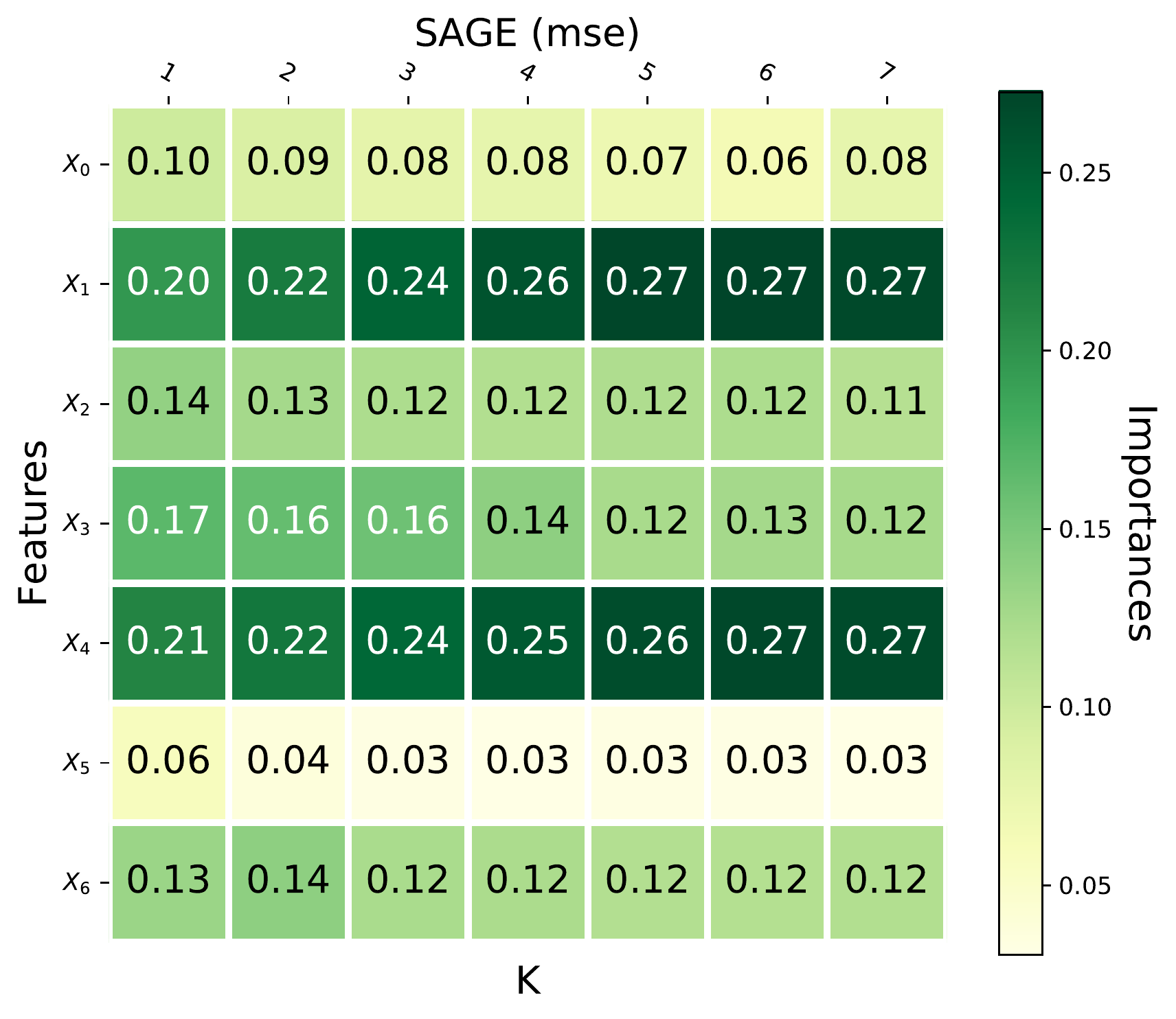}}\hfill
    \subfloat[mean(|TreeSHAP|) (interventional)]{\includegraphics[width=0.45\linewidth]{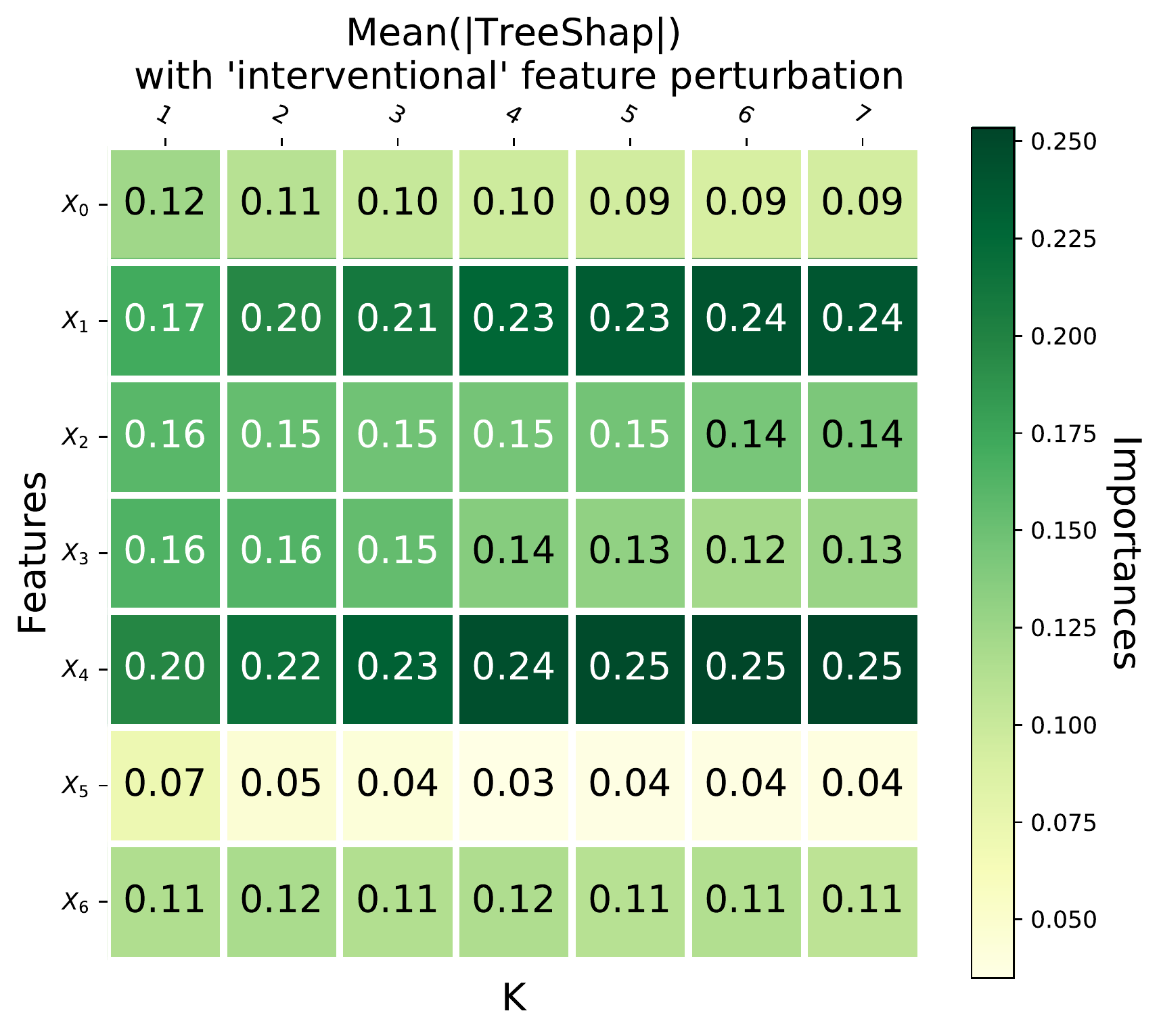}}\hfill
    \subfloat[mean($|$TreeSHAP$|$) (tree\_path\_dependent)]{\includegraphics[width=0.45\linewidth]{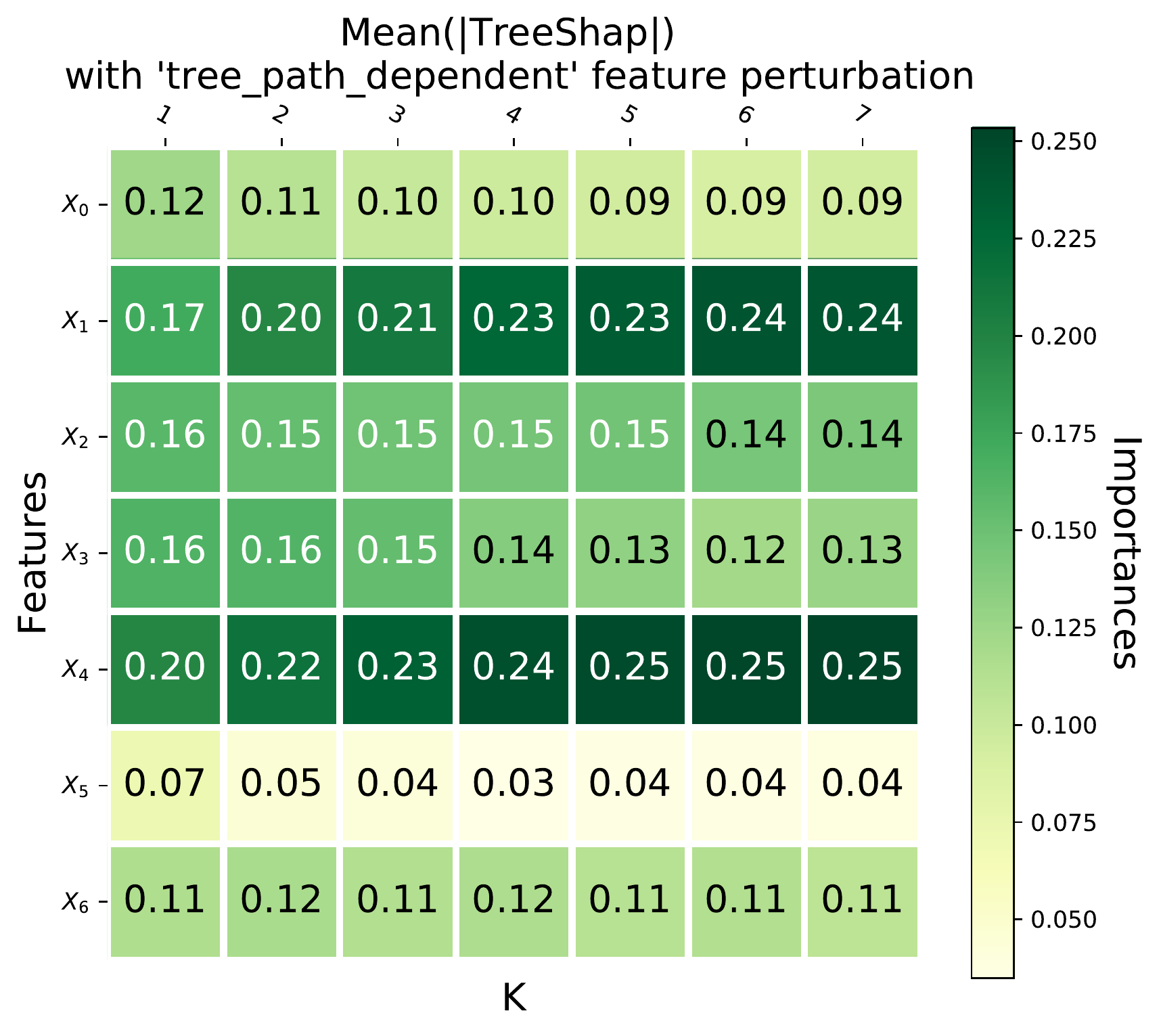}}\hfill
    \caption{Normalized importance scores derived from an ensemble of totally and non-totally randomized Extra-Trees (with $K=1,\dots,p$) for the global MDI importance measure and SAGE.}\hfill
    \label{app:fig:global}
\end{figure*}

Figure~\ref{app:fig:global} reports normalized importance scores derived from an ensemble of trees with increasing $K$ (\ie, the randomization parameter) on the \texttt{Led} dataset for SAGE with the mean squared loss (mse) as loss function, and the mean of the absolute value of TreeSHAP (for the predicted class) with both available parameters for feature perturbations. It appears that all three methods reflect the impact of $K$ on the measured importance scores similarly as the global MDI and SAGE using the cross-entropy loss function (Figure~\ref{fig:Kp:MDI+SAGE}). 

\section{Supplementary results for local importance measures}\label{app:supplementaryresults_local}
This section presents additional results and experiments that compare local importance measures.

\subsection{Local importances for \texttt{led} and \texttt{digits}}

\begin{figure}[ht!]
\centering
\subfloat[\textsc{led} \label{app:fig:exp:local:left}]{\includegraphics[width=0.48\linewidth]{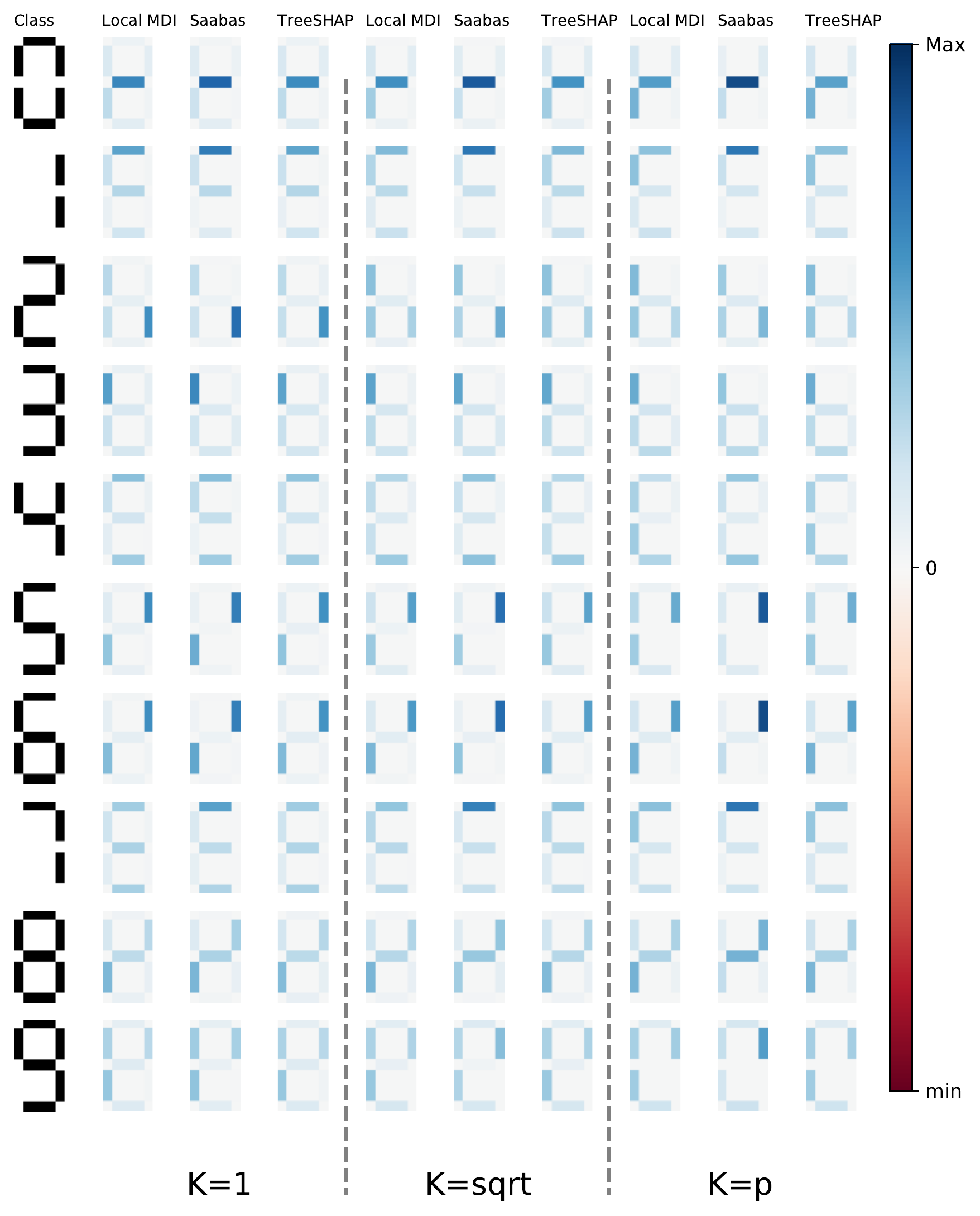}}
\subfloat[\textsc{digits}\label{app:fig:exp:local:right}]{\includegraphics[width=0.48\linewidth]{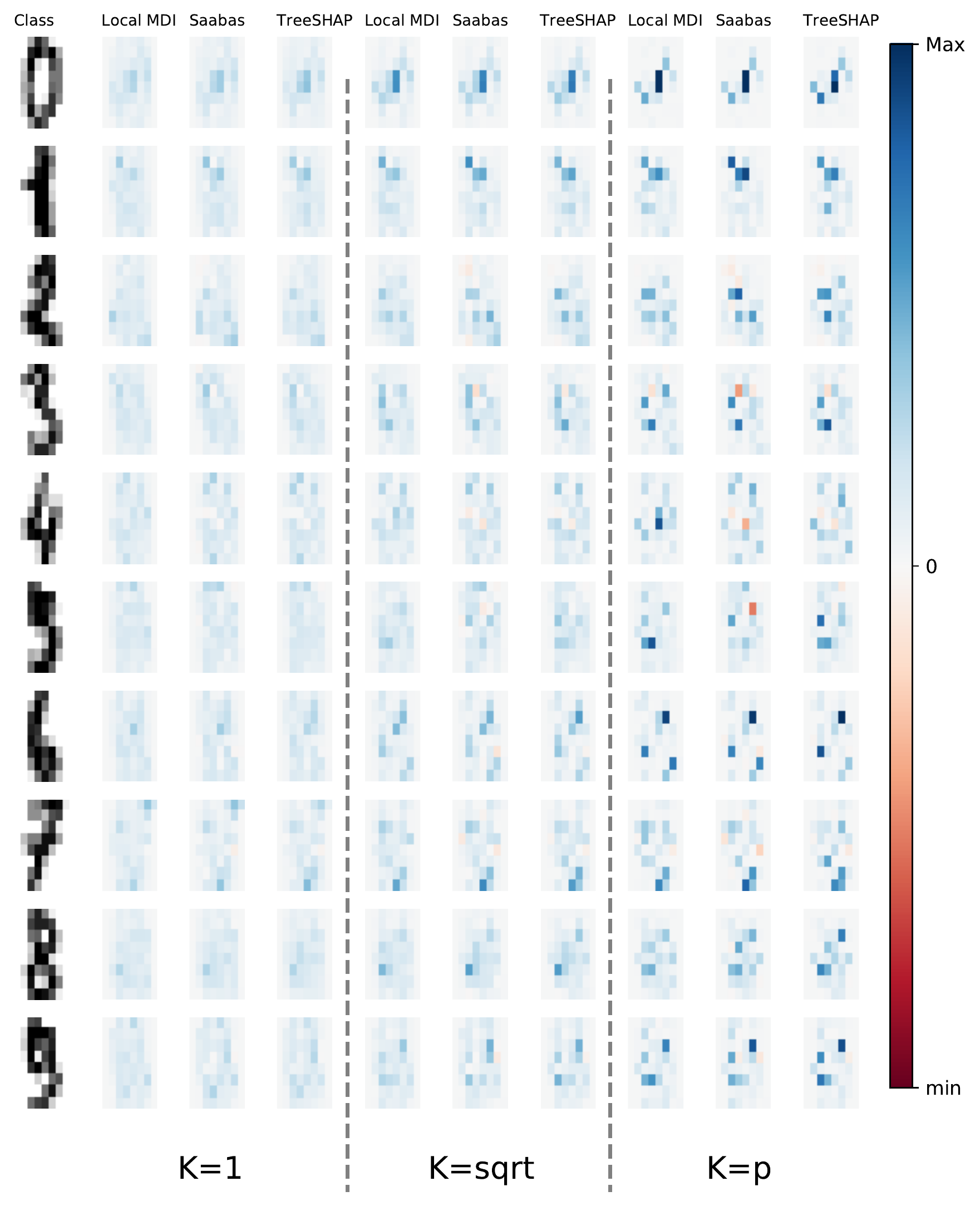}}
\caption{Local importances derived by local measures from a forest of 1000 Extra-Trees with $K \in \{1, \sqrt{p}, p\}$) for led (left) and digits (right).}

\label{fig:app:exp:local:K}
\end{figure}

\begin{figure}[htbp]
    \centering
    \includegraphics[width=0.4\linewidth]{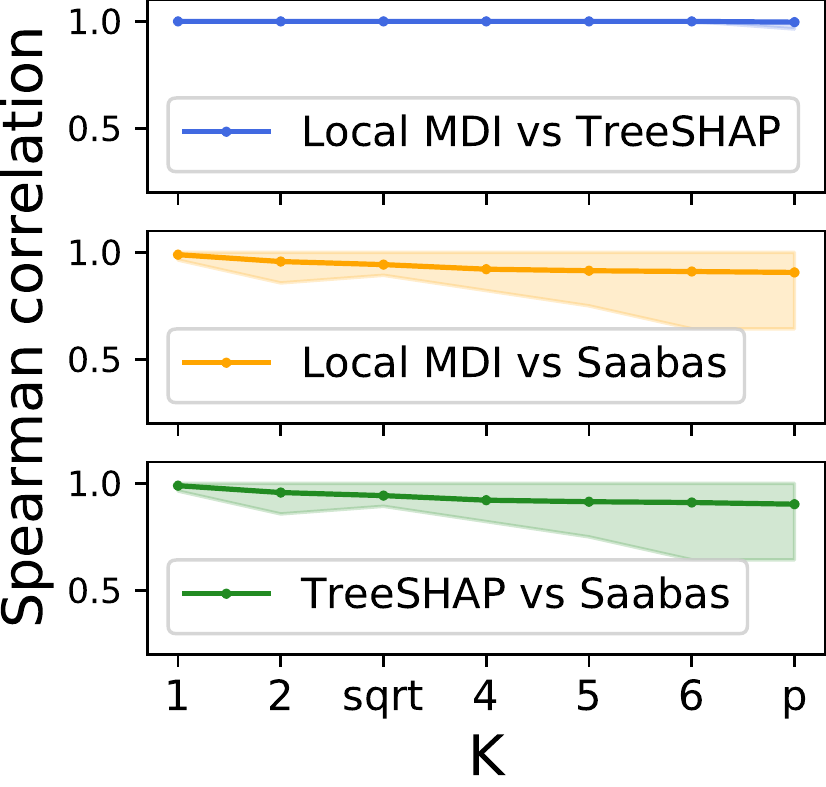}
    \caption{Mean correlation (over the samples)  w.r.t. increasing $K$ for between absolute importance scores for \texttt{led} ([min,max] is shaded). }
    \label{app:fig:local:led:corr+K}
\end{figure}

Figure \ref{app:fig:local:led:corr+K} shows the correlation between the (absolute value) of local importance scores of all pairs of methods for several values of the randomization parameter $K$.  Figures \ref{fig:app:exp:local:K} reports the local importances for three values of $K$ for \texttt{led} (left) and \texttt{digits} (right), showing more samples (all samples of \texttt{led} and one of each class for \texttt{digits}) than Figure~\ref{fig:exp:local}.

\subsection{Local importances on other datasets}

In addition to the \texttt{led} and \texttt{digits} datasets already used in Section~\ref{sec:illustration}, we consider here a few additional classification datasets. 
\texttt{led (sampled)} is a variant of the foretold problem where the learning set is made of (200) samples randomly drawn from the data distribution. The remaining datasets have been chosen from Scikit-learn datasets to cover mixed settings and differ by their dimensionality and their feature types {(both discrete and continuous)}. Table~\ref{app:tab:datasets_classification} summarizes their characteristics.

\begin{table}[hbtp]
    \caption{Datasets in classification}

    \centering
    \begin{tabular}{c|c | c|c }
       Name  & \shortstack{Nb of\\ samples} & \shortstack{Nb of\\features} & \shortstack{Feature\\types}\\\hline
       \texttt{Led} & 10 & 7 &Binary\\\hdashline
       \shortstack{\texttt{Led (sampled)}} & 200 & 7 & Binary\\\hdashline
       \texttt{Digits} & 1797 & 64 & Integer\\\hdashline
       \texttt{Iris} & 150 & 4 & Real \\\hdashline
       \texttt{Wine} & 178 & 13 & Integer, Real\\\hdashline
       \texttt{Breast cancer} & 569 & 30 & Real \\
    \end{tabular}
    \label{app:tab:datasets_classification}
\end{table}

 For every dataset, local feature importances are derived from a forest of 1000 totally randomized Extra-Trees ($K=1$). As in Section~\ref{sec:illustration}, Saabas and TreeSHAP are computed with respect to the predicted class. 
 
 For each sample, Pearson and Spearman correlations are computed between  the local MDI and Saabas importances and between the local MDI and treeSHAP importances. 
 
\begin{figure}[htbp]
     \centering
     \subfloat[\texttt{Led}]{\includegraphics[width=0.5\linewidth]{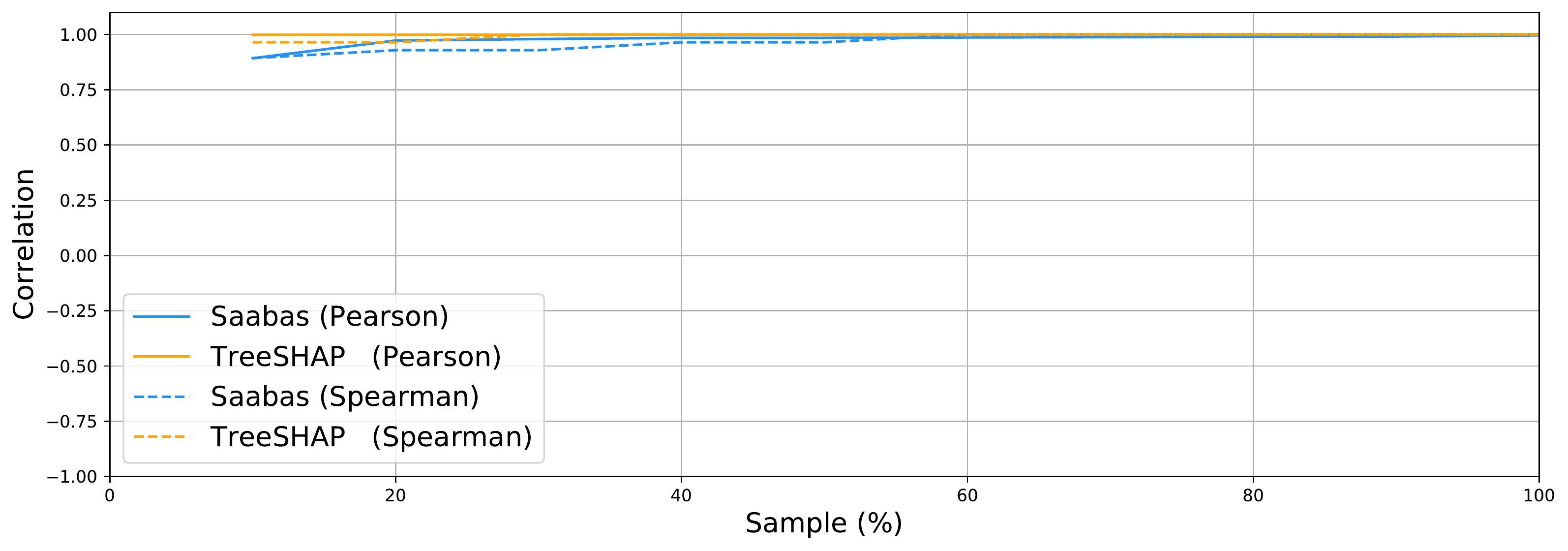}}
     \subfloat[\texttt{Led (sampled)}]{\includegraphics[width=0.5\linewidth]{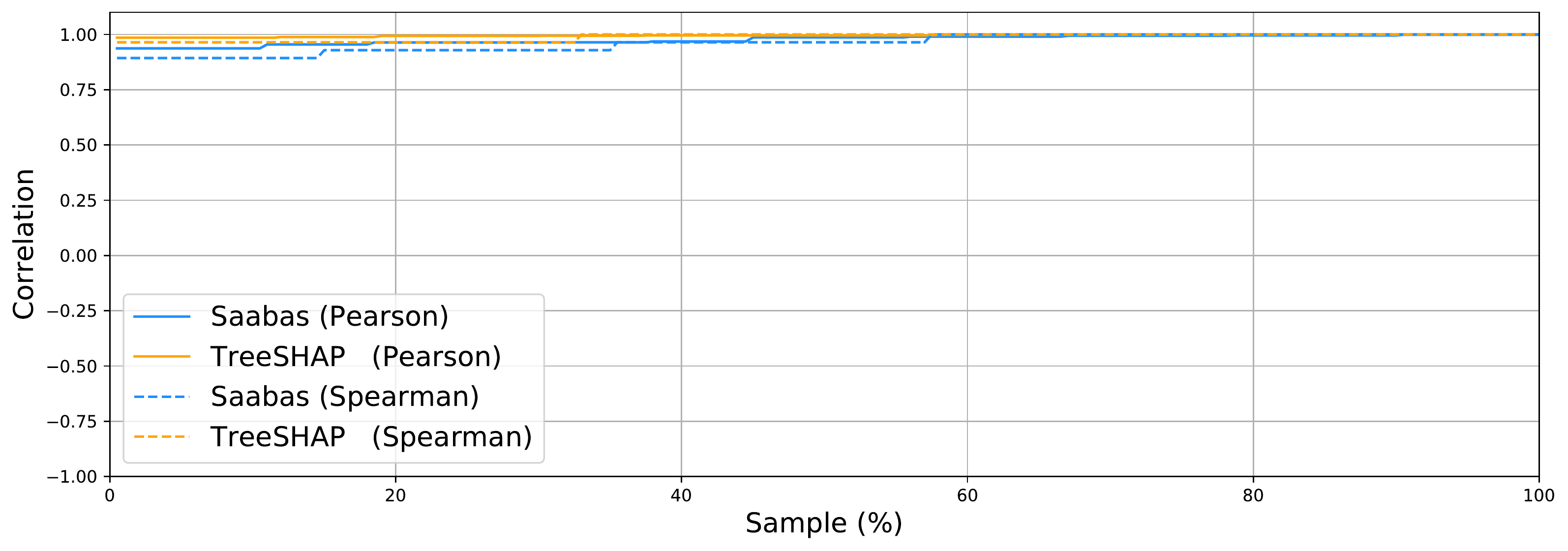}}
     
     \subfloat[\texttt{Iris}]{\includegraphics[width=0.5\linewidth]{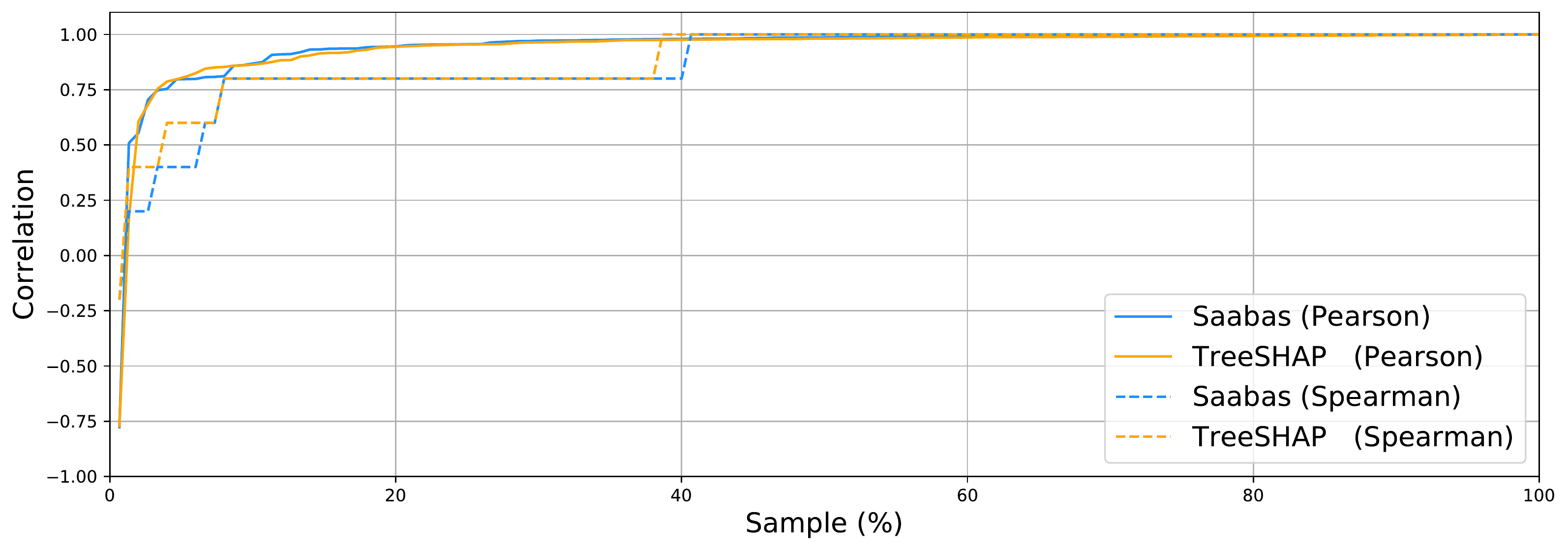}}
     \subfloat[\texttt{Wine}]{\includegraphics[width=0.5\linewidth]{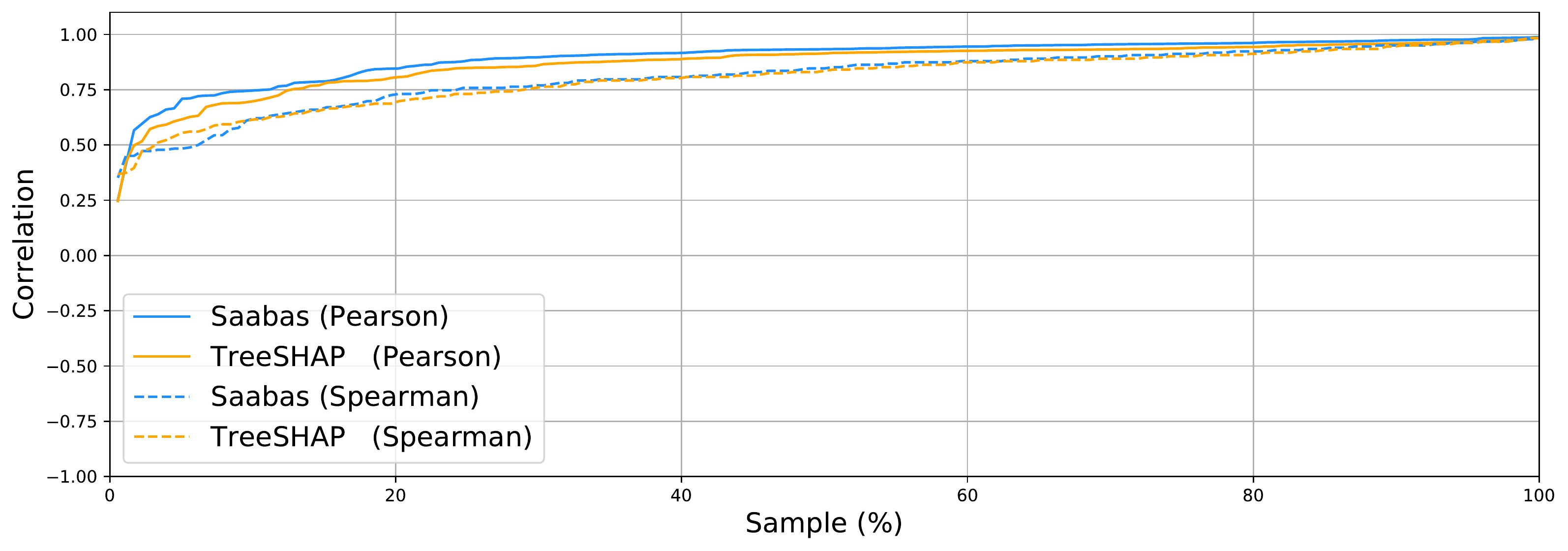}}
     
     \subfloat[\texttt{Breast cancer}]{\includegraphics[width=0.5\linewidth]{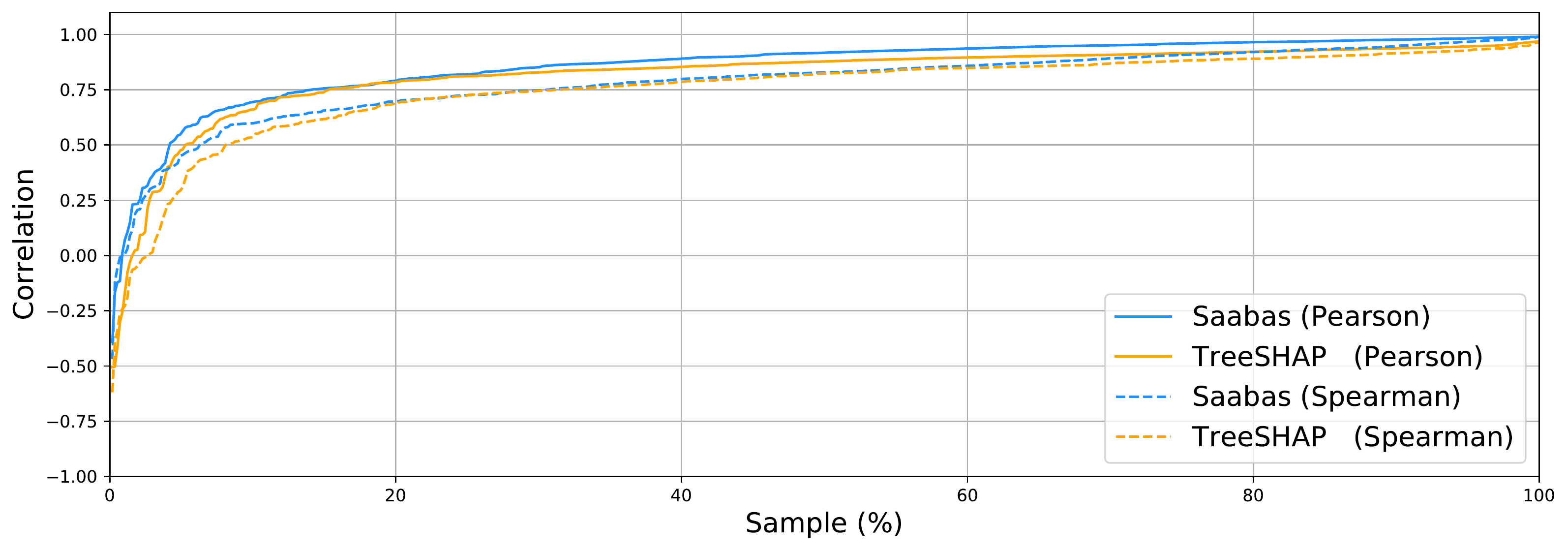}}
     \subfloat[\texttt{Digits}]{\includegraphics[width=0.5\linewidth]{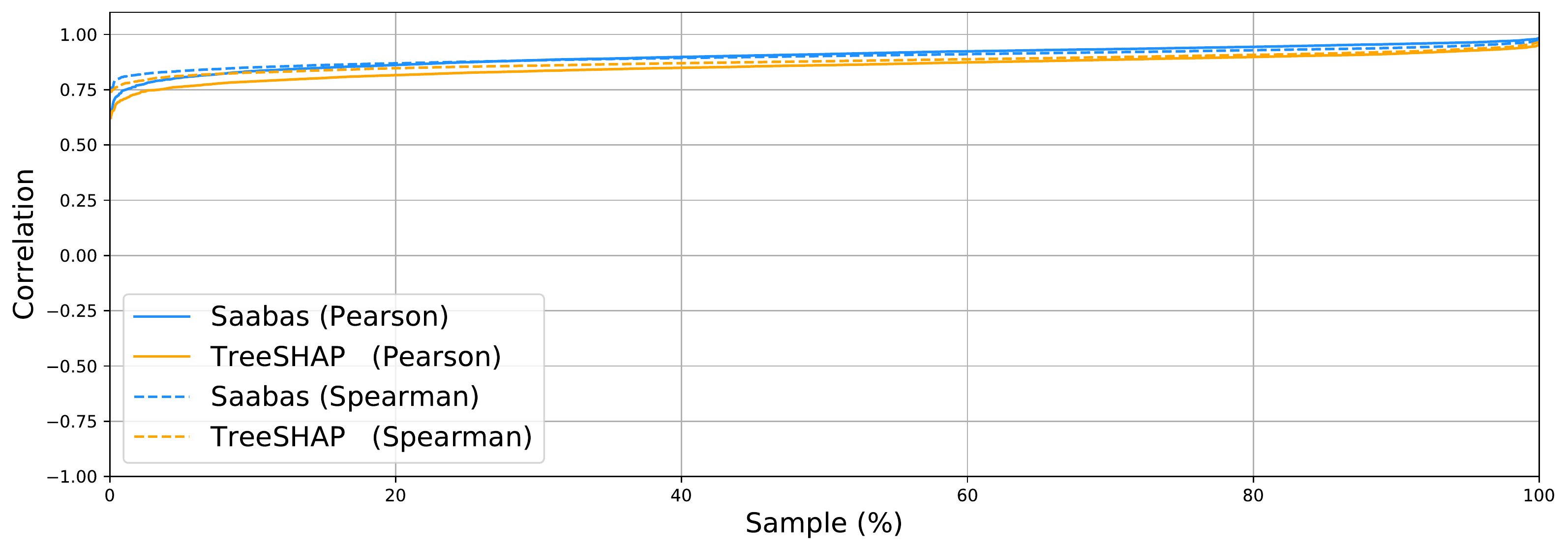}}
     
     \caption{Correlations between feature importance vectors across the samples ordered by increasing values.}
     \label{app:fig:classification_results}
 \end{figure}

 Figure~\ref{app:fig:classification_results} shows the correlations across all samples that are reordered by increasing correlation values. Table~\ref{app:tab:classification_results} summarizes these results.

 It can be observed that a large fraction of samples are associated to correlations close to $1$ suggesting that most of importance vectors are similar. More than $90\%$ importance vectors are similar (correlation $\ge 0.75$) in four datasets (out of six) and more than $80\%$ in two datasets. Note that this also shows that a fraction of samples (roughly $1\%$ to $10\%$ depending on the dataset) are associated with uncorrelated importance scores. This may come from the importance scores with respect to the predicted class (used in TreeSHAP and Saabas) that differs from the importance scores over all classes (local MDI). More experiments should be carried out to better understand the origin of these differences.

 \begin{table*}[htbp]
      \caption{Summary of results on classification datasets.}

     \centering
     \begin{tabular}{c|c|c|c|c}\hline
     \multicolumn{5}{c}{Saabas}\\\hline
     \multirow{2}{*}{Dataset} & \multirow{2}{*}{Correlation}& \multirow{2}{*}{Avg. corr. ($\pm$std)}& \multicolumn{2}{c}{Fraction of samples}\\
     &&&w/ correlation $\ge0.9$&w/ correlation $\ge0.75$\\\hline
     \multirow{2}{*}{\texttt{Wine}} & Pearson & $0.906 \;(\pm 0.101)$ & $72.47\%$ & $91.57\%$\\\cdashline{2-5}
     & Spearman & $0.843\;(\pm 0.128)$ & $44.38\%$ & $80.90\%$\\\hline
     \multirow{2}{*}{\texttt{Iris}} & Pearson & $0.949 \;(\pm 0.156)$ & $89.33\%$ & $95.33\%$\\\cdashline{2-5}
     & Spearman & $0.892\;(\pm 0.231)$ & $67.33\%$ & $91.33\%$\\\hline
     \multirow{2}{*}{\texttt{Breast cancer}} & Pearson & $0.899 \;(\pm 0.220)$ & $79.96\%$ & $91.04\%$\\\cdashline{2-5}
     & Spearman & $0.857\;(\pm 0.255)$ & $68.19\%$ & $84.71\%$\\   
     \hline
     \multirow{2}{*}{\texttt{Led}} & Pearson & $0.980 \;(\pm 0.025)$ & $100.00\%$ & $100.00\%$\\\cdashline{2-5}
     & Spearman & $0.989\;(\pm 0.016)$ & $100.00\%$ & $100.00\%$\\ \hline
     \multirow{2}{*}{\texttt{Led (sampled)}} & Pearson & $0.970 \;(\pm 0.034)$ & $100.00\%$ & $87.00\%$\\\cdashline{2-5}
     & Spearman & $0.978\;(\pm 0.017)$ & $100.00\%$ & $100.00\%$\\\hline
     \multirow{2}{*}{\texttt{Digits}} & Pearson & $0.915 \;(\pm 0.045)$ & $69.84\%$ & $99.44\%$\\\cdashline{2-5}
     & Spearman & $0.899\;(\pm 0.045)$ & $55.65\%$ & $99.50\%$\\     
     \end{tabular}
     \begin{tabular}{c|c|c|c|c}\hline\hline
     \multicolumn{5}{c}{TreeSHAP}\\\hline
     \multirow{2}{*}{Dataset} & \multirow{2}{*}{Correlation}& \multirow{2}{*}{Avg. corr. ($\pm$std)}& \multicolumn{2}{c}{Fraction of samples}\\
     &&&w/ correlation $\ge0.9$&w/ correlation $\ge0.75$\\\hline
     \multirow{2}{*}{\texttt{Wine}} & Pearson & $0.900 \;(\pm 0.104)$ & $70.22\%$ & $90.45\%$\\\cdashline{2-5}
     & Spearman & $0.852\;(\pm 0.121)$ & $47.75\%$ & $82.02\%$\\\hline
     \multirow{2}{*}{\texttt{Iris}} & Pearson & $0.947 \;(\pm 0.150)$ & $88.00\%$ & $96.67\%$\\\cdashline{2-5}
     & Spearman & $0.881\;(\pm 0.186)$ & $56.67\%$ & $92.00\%$\\  \hline
     \multirow{2}{*}{\texttt{Breast cancer}} & Pearson & $0.888 \;(\pm 0.223)$ & $79.26\%$ & $90.33\%$\\\cdashline{2-5}
     & Spearman & $0.841\;(\pm 0.254)$ & $62.57\%$ & $84.18\%$\\  \hline
     \multirow{2}{*}{\texttt{Led}} & Pearson & $1.000 \;(\pm 0.000)$ & $100.00\%$ & $100.00\%$\\\cdashline{2-5}
     & Spearman & $1.000\;(\pm 0.000)$ & $100.00\%$ & $100.00\%$\\ \hline \multirow{2}{*}{\texttt{Led (sampled)}} & Pearson & $0.990 \;(\pm 0.009)$ & $100.00\%$ & $100.00\%$\\\cdashline{2-5}
     & Spearman & $0.988\;(\pm 0.017)$ & $100.00\%$ & $100.00\%$\\\hline \multirow{2}{*}{\texttt{Digits}} & Pearson & $0.881 \;(\pm 0.047)$ & $38.01\%$ & $98.39\%$\\\cdashline{2-5}
     & Spearman & $0.891\;(\pm 0.041)$ & $44.80\%$ & $99.55\%$\\  
     \end{tabular}
     \label{app:tab:classification_results}
 \end{table*}

\section{Generalization to other impurity measures and to regression}\label{app:generalisation}

We have considered in most of our developments in the paper a categorical output $Y$ (\ie, a classification problem) and the use of Shannon entropy as impurity measure. \cite{louppe2013understanding} show that Equation \ref{eqn:popMDI} and the link between the irrelevance of $X_m$ and $Imp_\infty(X_m)$ remain valid for other impurity measures in classification, such as the Gini index, and can be extended to regression problems using variance as the impurity measure. Similarly, the local MDI measure can be extended to other impurity measures and thus in particular also to regression problems (\ie, a numerical output $Y$). Definition \ref{eqn:def:localmdi} is indeed generic and valid whatever the function $i(Y|t)\ge 0$ that measures the impurity of the output $Y$ at a tree node $t$. The link between local and global MDI as expressed in Equation \ref{eq:oilocalmdidecomp} (main paper) is also generic. Asymptotic results in Sections \ref{sec:asymptoticlocal} and \ref{sec:localshapley}, i.e., the decomposition in Theorem \ref{THM:LOCAL} and the link with Shapley value, remain valid with entropy $H$ replaced by the population version of the choosen impurity measure, denoted $I_\infty$ in what follows. Results in Section \ref{sec:localandrelevance} requires to redefine the notion of local irrelevance from the impurity function directly. If one defines a variable $X_m$ as {\it locally irrelevant} at $\bmm{x}$ w.r.t. $Y$ iff $I_\infty(Y|X_m=\bmm{x}_m,B=\bmm{x}_B)=I_\infty(Y|B=\bmm{x}_B)$ for all $B\subseteq V^{-m}$ \footnote{This definition matches Definition \ref{def:localirrelevance} when $I_\infty$ is the entropy $H$.}, then Theorem \ref{COR:=0} still applies.

As an illustration, if one considers regression using the empirical variance as the impurity measure:
$$
i(Y|t) = \frac{1}{N_t}\sum_{i\in t} (y_i-\frac{1}{N_t} \sum_{i\in t} y_i)^2,
$$ where $N_t$ is the number of instances in node $t$ and $y_i$ are
their output values, then $I_\infty$ is the (conditional) population variance and the decomposition in Equation \ref{THM:LOCAL:EQN} (main paper) becomes:
\begin{equation}
Imp_\infty(X_m,\bmm{x}) = \sum_{k=0}^{p-1} \dfrac{1}{C^k_p} \dfrac{1}{p-k} \sum_{B\in \mathcal{P}(V^{-m})} \mbox{Var}(Y|B=\bmm{x}_B)- \mbox{Var}(Y|B=\bmm{x}_B, X_m= x_m),
\end{equation}
where $\mbox{Var}(Y|B=\bmm{x}_B)$ is the conditional variance of the output:
\begin{equation}
\mbox{Var}(Y|B=\bmm{x}_B) = E_{Y|B=\bmm{x}_B}\{(Y-E_{Y|B=\bmm{x}_B}\{Y\})^2\}.
\end{equation}
Global and local MDI importances of totally randomized trees are in this case Shapley values respectively with respect to the following characteristic functions:
\begin{eqnarray}
  v(S) &=& \mbox{Var}(Y)-E_S\{\mbox{Var}(Y|S)\}\\
  v_{loc}(S;\bmm{x}) &= & \mbox{Var}(Y)- \mbox{Var}(Y|S=\bmm{x}_S),
\end{eqnarray}
and the decomposition in Equation \ref{eq:localdecompinfo} therefore becomes:
\begin{equation}
\mbox{Var}(Y)-E_S\{\mbox{Var}(Y|X)\} =
\sum_{m=1}^p\sum_{\bmm{x}\in\mathcal{V}} P(V=\bmm{x})
Imp_\infty(X_m,\bmm{x}).
\end{equation}
Finally, locally irrelevant variables (null players) are such that $\mbox{Var}(Y|X_m=\bmm{x}_m,B=\bmm{x}_B)=\mbox{Var}(Y|B=\bmm{x}_B)$ for all $B\in V^{-m}$.

\section{Notations, and definitions of entropies and mutual information}

{As a minimal introduction to information theory, we recall in this section several definitions from information theory (see \cite{cover2012elements}, for further properties). The presentation below is largely based on the Supplementary material of \citep{louppe2013understanding}, which is reproduced here for the convenience of the reader.

We suppose that we are given a probability space $(\Omega, {\cal E}, \mathbb{P})$ and  consider random variables
defined on it taking a finite number of possible values. We use upper case letters to denote such random variables (e.g. $X, Y, Z, W \ldots$)  and calligraphic letters (e.g. $\cal X, Y, Z, W \ldots$) to denote their image sets (of finite cardinality), and lower case letters (e.g. $x, y, z, w \ldots$) to denote one of their possible values.
For a (finite) set of (finite) random variables $ X = \{X_{1}, \ldots , X_{i}\}$, we denote by $P_{X}(x) = P_{X}(x_{1}, \ldots , x_{i})$ the probability $\mathbb{P}(\{ \omega \in \Omega \mid  \forall \ell : 1, \ldots, i: X_{\ell}(\omega) =x_{\ell}\})$, and by ${\cal X} = {\cal X}_{1} \times \cdots \times {\cal X}_{i}$ the set of joint configurations of these random variables. Given two sets of random variables, $X = \{X_{1}, \ldots , X_{i}\}$ and $Y=\{Y_{1}, \ldots , Y_{j}\}$, we denote by $P_{X \mid Y}(x \mid y) = {P_{X, Y} (x,  y)}/ {P_{Y}(y)}$ the conditional density of $X$ with respect to $Y$.\footnote{To avoid problems, we suppose that all probabilities are strictly positive, without fundamental limitation.}

With these notations, the joint (Shannon) entropy of a set of random variables $X =\{X_{1}, \ldots , X_{i}\}$ is thus defined by
\begin{equation*}
H(X)  = - \sum_{x \in {\cal X}}P_{X} (x)\log_{2}P_{X }(x),
\end{equation*}
while the mean conditional entropy of a set of random variables $X = \{X_{1}, \ldots , X_{i}\}$, given the values of another set of random variables $Y=\{Y_{1}, \ldots , Y_{j}\}$ is defined by
\begin{equation*}
H(X\mid Y) = - \sum_{x \in {\cal X}} \sum_{y \in {\cal Y}} P_{X, Y} (x, y) \log_{2} P_{X \mid Y} (x  \mid y).
 \end{equation*}
The mutual information among the set of random variables $X =\{X_{1}, \ldots , X_{i}\}$ and the set of random variables $Y=\{Y_{1}, \ldots , Y_{j}\}$ is defined by
 \begin{eqnarray*}
 I(X; Y) & = &- \sum_{x \in {\cal X}} \sum_{y \in {\cal Y}} P_{X, Y} (x, y) \log_{2} \frac{P_{X}(x) P_{Y}(y)}{P_{X,Y}(x,y)} \\
 & = & H(X) - H(X \mid Y) \\
 & = &  H(Y) - H(Y \mid X).
 \end{eqnarray*}
 The mean conditional mutual information among the set of random variables $X =\{X_{1}, \ldots , X_{k}\}$ and the set of random variables $Y=\{Y_{1}, \ldots , Y_{j}\}$, given the values of a third set of random variables $Z=\{Z_{1}, \ldots , Z_{i}\}$, is defined by
 \begin{eqnarray*}
 I(X; Y \mid Z) &= &H(X \mid Z) - H(X \mid Y, Z)\\
 & = & H(Y \mid Z) - H(Y \mid X, Z)\\
& = & - \sum_{x \in {\cal X}} \sum_{y \in {\cal Y}} \sum_{z \in {\cal Z}} P_{X, Y, Z} (x, y, z) \log_{2} \frac{P_{X \mid Z}(x \mid z) P_{Y\mid Z}(y \mid z)}{P_{X,Y \mid Z}(x,y \mid z)} .
\end{eqnarray*}

We also recall the chaining rule
\begin{equation*}
I(X, Z ; Y \mid W ) = I(X; Y \mid W  ) + I( Z ; Y \mid W, X),
\end{equation*}
and the symmetry of the (conditional) mutual information among sets of random variables
\begin{equation*}
I(X ; Y \mid Z) = I(Y ;  X  \mid Z).
\end{equation*}
}



\end{document}